\documentclass[british,10pt,twocolumn,letterpaper]{article}
\usepackage[T1]{fontenc}
\usepackage[utf8]{luainputenc}
\usepackage{color}
\usepackage{wrapfig}
\usepackage{units}
\usepackage{amsmath}
\usepackage{graphicx}

\makeatletter

\providecommand{\tabularnewline}{\\}

\usepackage{iccv}
\usepackage{times}
\usepackage{epsfig}
\usepackage{graphicx}
\usepackage{amsmath}
\usepackage{amssymb}

\usepackage{colortbl}
\usepackage{arydshln}
\usepackage{makecell, pict2e}

\usepackage{rotating}

\usepackage{color}

\usepackage[french,english]{babel}

\usepackage{gensymb}

\definecolor{gray}{gray}{0.65}
\definecolor{lightgray}{gray}{0.8}
\definecolor{pink}{RGB}{219, 48, 122}

\usepackage[pagebackref=false,breaklinks=true,letterpaper=true,colorlinks,bookmarks=false]{hyperref}

 \iccvfinalcopy 

\def\iccvPaperID{1460} 

\ificcvfinal\fi

\@ifundefined{showcaptionsetup}{}{%
 \PassOptionsToPackage{caption=false}{subfig}}
\usepackage{subfig}
\makeatother

\usepackage{babel}
\begin{document}

\title{\vspace{-2em}
Person Recognition in Personal Photo Collections}

\author{\vspace{-2em}
}

\author{Seong Joon Oh\qquad{}Rodrigo Benenson\qquad{}Mario Fritz\qquad{}Bernt
Schiele\vspace{0.5em}
\\
\begin{tabular}{c}
Max Planck Institute for Informatics\tabularnewline
Saarbr\"ucken, Germany\tabularnewline
\texttt{\small{}\{joon,benenson,mfritz,schiele\}@mpi-inf.mpg.de}\tabularnewline
\end{tabular}\vspace{0em}
}
\maketitle
\begin{abstract}
Recognising persons in everyday photos presents major challenges
(occluded faces, different clothing, locations, etc.) for machine
vision. We propose a convnet based person recognition system on which
we provide an in-depth analysis of informativeness of different body
cues, impact of training data, and the common failure modes of the
system. In addition, we discuss the limitations of existing benchmarks
and propose more challenging ones. Our method is simple and is built
on open source and open data, yet it improves the state of the art
results on a large dataset of social media photos (PIPA).

\end{abstract}
\makeatletter 
\renewcommand{\paragraph}{%
\@startsection{paragraph}{4}%
{\z@}{1.0ex \@plus 1ex \@minus .2ex}{-1em}%
{\normalfont \normalsize \bfseries}%
}
\makeatother\setlength{\textfloatsep}{1.5em}

\section{\label{sec:Introduction}Introduction}

\begin{figure}
\begin{centering}
\vspace{-1em}
\arrayrulecolor{gray}
\par\end{centering}

\begin{centering}
\includegraphics[bb=0bp 0bp 601bp 600bp,width=0.25\columnwidth,height=0.25\columnwidth]{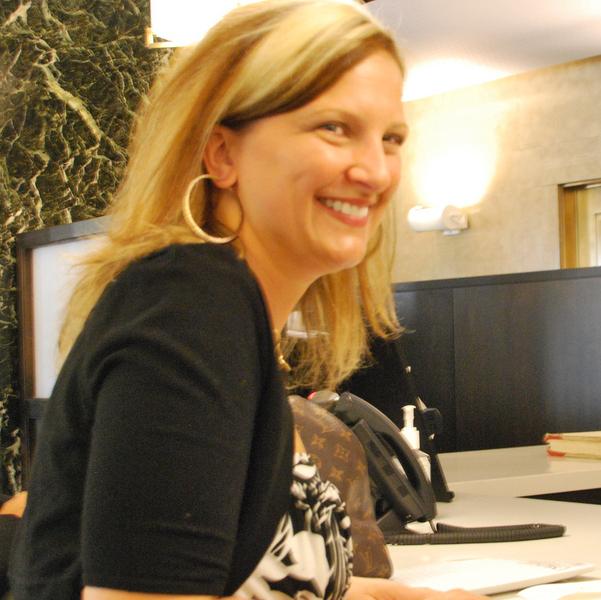}\includegraphics[bb=0bp 0bp 721bp 721bp,width=0.25\columnwidth,height=0.25\columnwidth]{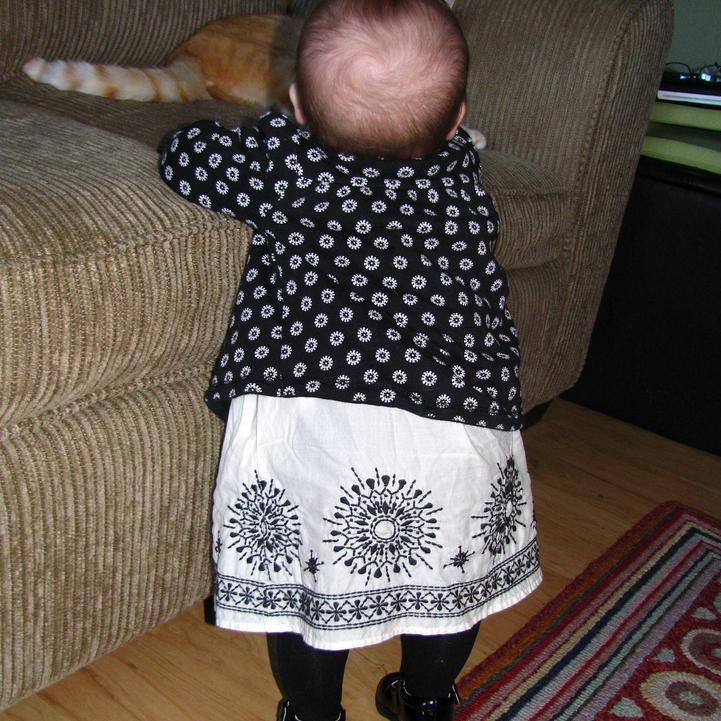}\includegraphics[bb=0bp 0bp 501bp 501bp,width=0.25\columnwidth,height=0.25\columnwidth]{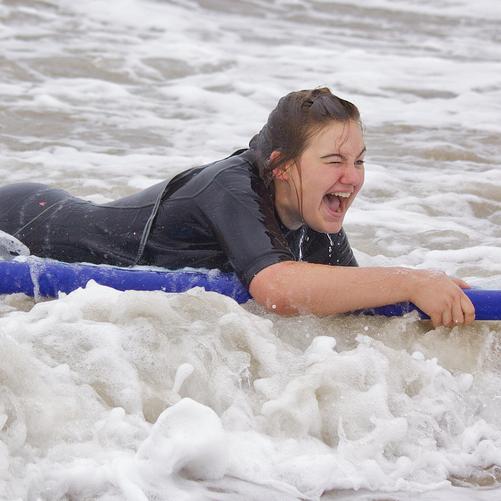}\includegraphics[bb=0bp 0bp 401bp 401bp,width=0.25\columnwidth,height=0.25\columnwidth]{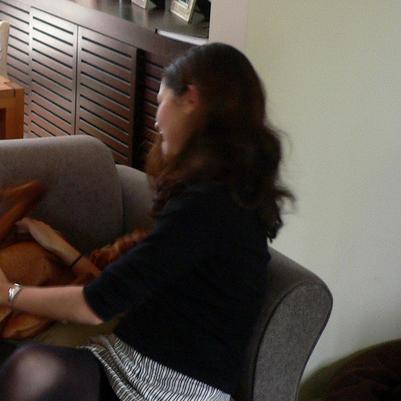}
\par\end{centering}

\vspace{0.5em}

\begin{raggedright}
\begin{tabular}{lcc|ccc|ccc|ccc}
\hspace*{-0.6em}Head & \hspace*{-0.6em}\includegraphics[height=0.8em]{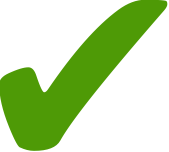} & \hspace*{-0.17em} & \hspace{0.8em} & \includegraphics[height=1em]{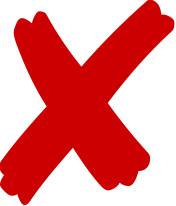} & \hspace{0.6em} & \hspace{0.9em} & \includegraphics[height=1em]{figures_arxiv/figure1f} & \hspace{0.49em} & \hspace{0.95em} & \includegraphics[height=1em]{figures_arxiv/figure1f} & \hspace{10em}\tabularnewline
\hspace*{-0.6em}Body & \hspace*{-0.6em}\includegraphics[height=0.8em]{figures_arxiv/figure1e} & \hspace*{-0.17em} &  & \includegraphics[height=0.8em]{figures_arxiv/figure1e} &  &  & \includegraphics[height=1em]{figures_arxiv/figure1f} &  &  & \includegraphics[height=1em]{figures_arxiv/figure1f} & \tabularnewline
\hspace*{-0.6em}{\scriptsize{}Attributes} & \hspace*{-0.6em}\includegraphics[height=0.8em]{figures_arxiv/figure1e} & \hspace*{-0.17em} &  & \includegraphics[height=0.8em]{figures_arxiv/figure1e} &  &  & \includegraphics[height=0.8em]{figures_arxiv/figure1e} &  &  & \includegraphics[height=1em]{figures_arxiv/figure1f} & \tabularnewline
\hspace*{-0.6em}{\small{}All cues} & \hspace*{-0.6em}\includegraphics[height=0.8em]{figures_arxiv/figure1e} & \hspace*{-0.17em} &  & \includegraphics[height=0.8em]{figures_arxiv/figure1e} &  &  & \includegraphics[height=0.8em]{figures_arxiv/figure1e} &  &  & \includegraphics[height=0.8em]{figures_arxiv/figure1e} & \tabularnewline
\end{tabular}
\par\end{raggedright}

\arrayrulecolor{black}\vspace{0.5em}

\protect\caption{\label{fig:teaser}Person recognition in photo albums is hard. To
handle the diverse scenarios we need to exploit multiple cues from
different body regions and information sources. Photos show test cases
successfully recognised by our system, ticks indicate which ingredient
could handle it. For example, the surfer is not recognised when using
only head or head+body cues. However, it is successfully recognised
when the additional attribute cues are provided.}
\end{figure}

Person recognition in private photo collections is challenging: people
can be shown in all kinds of poses and activities, from arbitrary
viewpoints including back views, and with diverse clothing (e.g. on
the beach, at parties, etc., see Figure \ref{fig:teaser}). This paper
presents an in-depth analysis of the problem of person recognition
in photo albums: given a few annotated training images of a person
(possibly from different albums), and a single image at test time,
can we tell if the image contains the same person? 

Intuitively, the ability to recognize faces in the wild \cite{Huang2007Lfw}
is an important ingredient. However, when persons are engaged in an
activity (i.e. not posing) their face becomes only partially visible
(non-frontal, occlusion) or simply fully non-visible (back-view).
Therefore, additional information is required to reliably recognize
people. We explore three other sources: first, body of a person contains
information about their shape and appearance; second, human attributes
such as gender and age help to reduce the search space; and third,
scene context further reduces ambiguities. 

The main contributions of the paper are the following. First, we provide
a detailed analysis of performance of different cues (\S\ref{sec:Cues-for-recognition}).
Second, we propose a more realistic and challenging experimental protocols
over PIPA (\S\ref{sub:PIPA-splits}) on which a deeper understanding
of robustness of different cues can be attained (\S\ref{sub:Importance-of-features}).
Third, in the process, we obtain best results on the recently proposed
PIPA dataset and show that previous performance can be matched without
specialized face recognition or pose estimation (\S\ref{sec:Test-set-results}).
Fourth, we analyse remaining failure modes (\S\ref{sec:failures-analysis}).
Additionally, our top-performing method is based only on open source
code and data, and the new experimental setups (\S\ref{sub:PIPA-splits}),
trained models, results, and attribute annotations are available at
\textcolor{pink}{http://goo.gl/DKuhlY}. 

\subsection{\label{sub:Related-work}Related work}

\paragraph{Data type}

The bulk of previous work on person recognition focus either on facial
features \cite{Huang2007Lfw} (only the head/face is visible), or
on the surveillance scenario \cite{Benfold2009BmvcTownCentre,Bedagkar2014IvcPersonReIdSurvey}
(full body is visible, usually in low resolution). Both settings have
seen a recent shift from sophisticated classifiers based on hand-crafted
features and metric learning approaches \cite{Guillaumin2009Iccv,Chen2013CvprBlessing,Cao2013IccvTransferLearning,Lu2014ArxivGaussianFace,Li2013Cvpr,Zhao2013IccvSalienceMatching,Bak2014WacvBrownian},
towards methods based on deep learning \cite{Taigman2014CvprDeepFace,Sun2014ArxivDeepId2plus,Zhou2015ArxivNaiveDeepFace,Schroff2015ArxivFaceNet,Li2014CvprDeepReID,Yi2014Arxiv,Hu2014Accvw}.

In this paper we tackle a different scenario, where persons may appear
at different zoom levels (e.g. only head, upper torso, full body visible),
and in any pose (e.g. sitting, running, posing), and from any point
of view (e.g. front, side, back view), see Figures \ref{fig:teaser}
and \ref{fig:success-O-split}. The ``Gallagher collection person
dataset'' \cite{Gallagher2008Cvpr} was the first dataset covering
this scenario; however, it is quite small (\textasciitilde{}600 images,
32 identities) and only frontal faces are annotated. We build our
paper upon the recently introduced PIPA dataset \cite{Zhang2015CvprPiper}
which is two orders of magnitude larger (\textasciitilde{}40k images,
\textasciitilde{}2k identities), more diverse, and also provides identity
annotations when the face is not visible. We describe PIPA in more
detail in \S \ref{sec:PIPA-dataset}.

\paragraph{Recognition tasks}

There exist multiple tasks related to person recognition \cite{Gong2014PersonReIdBook}
differing mainly in the amount of training and testing data. Face
and surveillance re-identification is most commonly done via ``verification''
(one reference image, one test image; do they show the same person?)
\cite{Huang2007Lfw,Bedagkar2014IvcPersonReIdSurvey}. The scenario
of our interest is \textasciitilde{}$20$ training images and one
test image.  

Other related tasks are, for instance, face clustering \cite{Cui2007ChiEasyAlbum,Schroff2015ArxivFaceNet},
finding important people \cite{Mathialagan2015Arxiv}, or associating
names in text to faces in images \cite{Everingham2006Bmvc,Everingham2009Ivc}.

\paragraph{Recognition cues}

The base cue for person recognition is the appearance of the face
itself. Face normalization (``frontalisation'') \cite{Zhu2013Iccv,Taigman2014CvprDeepFace,Ding2015Arxiv}
improves robustness to pose, view-point and illumination. Similarly,
pose-independent descriptors can be built for the body \cite{Cheng2011Bmvc,Gandhi2013Cvpr,Zhang2015CvprPiper}.

Multiple other cues have been explored, for example: attributes classification
\cite{Kumar2009Cvpr,Layne2012Bmvc}, explicit cloth modelling \cite{Gallagher2008Cvpr},
relative camera positions \cite{Garg2011Cvpr}, social context \cite{Gallagher2007Cvpr,Stone2008Cvprw},
space-time priors \cite{Lin2010Eccv}, and photo-album priors \cite{Shi2013Iccv}.

The PIPA dataset was introduced together with the reference PIPER
method \cite{Zhang2015CvprPiper}. PIPER obtains promising results
combining three ingredients: a convnet (AlexNet \cite{Krizhevsky2012Nips})
pre-trained on ImageNet \cite{Deng2009CvprImageNet}, the DeepFace
re-identification convnet (trained on a large private faces dataset)
\cite{Taigman2014CvprDeepFace}, and Poselets \cite{Bourdev2009IccvPoselets}
(trained on H3D) to obtain robustness to pose variance. In contrast,
this paper considers features based on open data and use the same
AlexNet network for all the image regions considered, thus providing
a direct comparison of contributions from different image regions.

\section{\label{sec:PIPA-dataset}PIPA dataset}

The recently introduced PIPA dataset (``People In Photo Albums'')
\cite{Zhang2015CvprPiper} is, to the best of our knowledge, the first
dataset to annotate identities of people with back views. The annotators
labelled many instances that can be considered hard even for humans
(Figure \ref{fig:success-O-split}). PIPA features $37\,107$ Flickr
personal photo album images (Creative Commons), with $63\,188$ head
bounding boxes of $2\,356$ identities. The dataset is partitioned
into train, validation, test, and leftover sets, with rough ratio
$45\negmedspace:\negmedspace15\negmedspace:\negmedspace20\negmedspace:\negmedspace20$.
Up to annotation errors, neither identities nor photo albums by the
same uploader are shared among these sets.

For valid comparisons, we follow the PIPA protocol in \cite{Zhang2015CvprPiper}.
The training set is used for feature learning and the validation set
for exploring and optimising options. The test set is for evaluation
of our methods (Table \ref{tab:test-set-accuracy}); it is itself
split in two parts, $\mbox{test}_{0}$/ $\mbox{test}_{1}$, with roughly
the same number of instances per identity. Given $\mbox{test}_{0}$
a classifier is learnt for each identity ($11$ examples per identity
on average), and these are evaluated on $\mbox{test}_{1}$ (and vice-versa).
Later we consider more challenging splits than the PIPA default (\S
\ref{sub:PIPA-splits}).

At test time, the system is fed with the photo of the test instance
and the ground truth head annotation (tight around the skull, face
and hair included; not fully visible heads are hallucinated by the
annotators). The task is to find the corresponding identity of the
head.

In the next section, various image regions and the corresponding recognition cues are defined (\S \ref{sub:Body-regions}), and their validation set performances are compared (\S \ref{sec:how-informative-body-region} to \S \ref{sec:Attributes}). 
The performance of our final system and comparisons to other methods and baselines are provided in \S \ref{sec:Test-set-results}.
\S \ref{sec:failures-analysis} will present an in-depth analysis of the systems, including the performance on the more realistic and challenging PIPA splits.

\section{\label{sec:Cues-for-recognition}Cues for recognition}

\begin{wrapfigure}{O}{0.43\columnwidth}%
\centering{}\vspace{-3.5em}
\includegraphics[height=0.6\columnwidth]{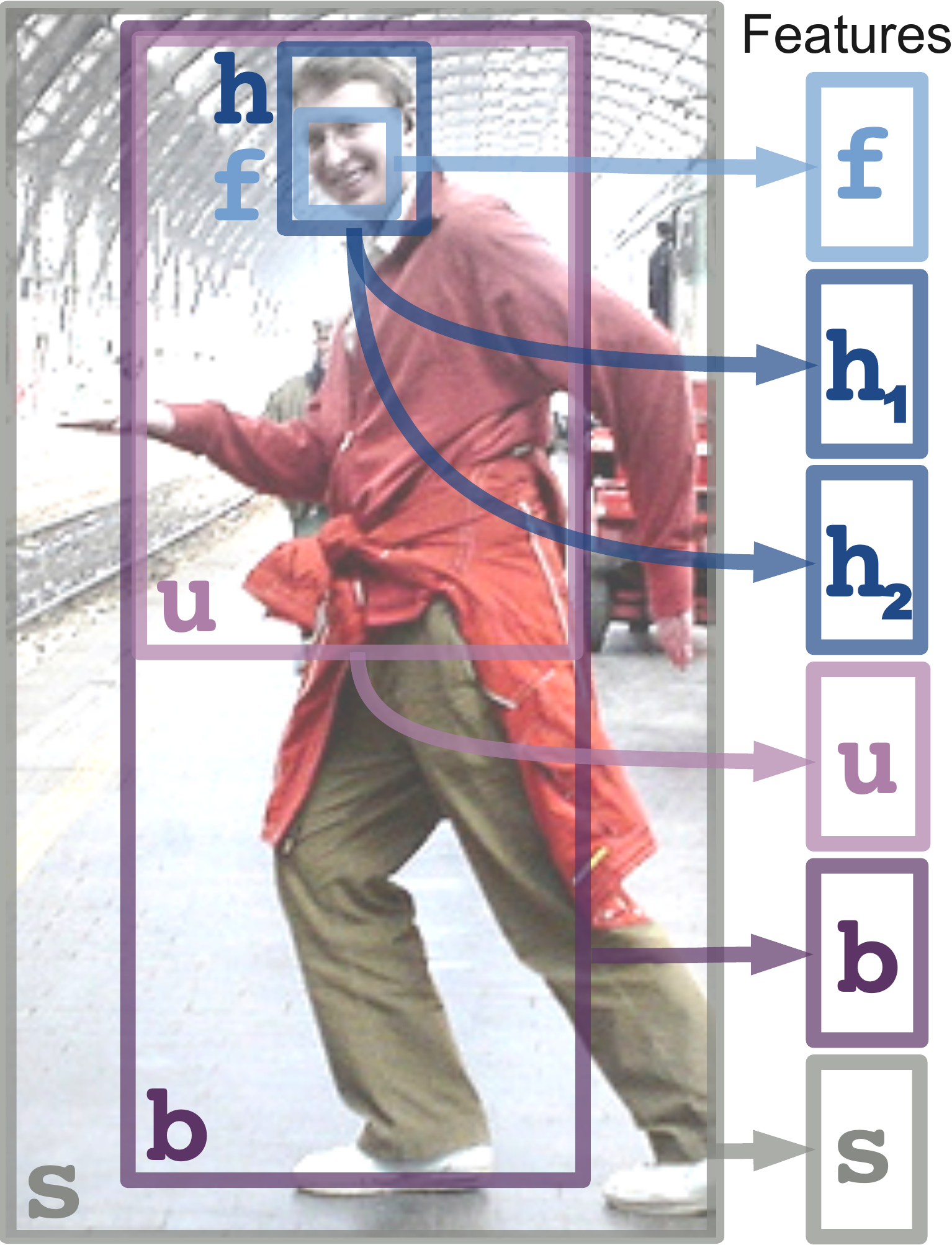}\protect\caption{\label{fig:image-regions}Regions considered for feature extraction:
face $\ensuremath{\texttt{f}}$, head $\ensuremath{\texttt{h}}$,
upper body $\ensuremath{\texttt{u}}$, full body $\ensuremath{\texttt{b}}$,
and scene $\ensuremath{\texttt{s}}$. More than one feature vector
can be extracted per region (e.g. $\ensuremath{\texttt{h}}_{1}$,
$\ensuremath{\texttt{h}}_{2}$ ).}
\vspace{-1em}
\end{wrapfigure}%

Our person recognition system is performant yet simple. At test time,
given a (ground truth) head bounding box, we estimate (based on the
box size) five different regions depicted. Each region is fed into
one (or more) convnet(s) to obtain a set of feature vectors. The vectors
are concatenated and fed into a linear SVM, trained per identity as
one versus the rest (on $\mbox{test}_{\nicefrac{0}{1}}$). In our
final system all features are computed using the seventh layer of
an AlexNet \cite{Krizhevsky2012Nips} pre-trained for ImageNet classification
(albeit we explore alternatives in the next sections). The cues only
differ among each other by the image region considered, and by the
fine-tuning used to alter the AlexNet model (type of data or surrogate
task).\cite{Krizhevsky2012Nips} 

Compared to PIPER \cite{Zhang2015CvprPiper}, we merge cues with a
simpler schema and do not use specialized face recognition or pose
estimation. Instead, we explore different directions: how informative
are fixed body regions (no pose estimation) (\S\ref{sec:how-informative-body-region})?
How much does scene context help (\S\textcolor{red}{\ref{sec:Scene}})?
And how much do we gain by using extended data (\S\textcolor{red}{\ref{sec:Additional-training-data}}
\& \S\textcolor{red}{\ref{sec:Attributes}})? This section is based
exclusively on validation set.

\subsection{\label{sub:Body-regions}Image regions used}

We choose five different image regions based on the ground truth head
annotation (given at test time, see \S\ref{sec:PIPA-dataset}). The
head rectangle $\ensuremath{\texttt{h}}$ corresponds to the ground
truth annotation. The full body rectangle $\ensuremath{\texttt{b}}$
is defined as $\left(3\negthinspace\times\negthinspace\mbox{head width},\right.$
$\left.6\negthinspace\times\negthinspace\mbox{head height}\right)$,
with the head at the top centre of the full body. The upper body rectangle
$\ensuremath{\texttt{u}}$ is the upper-half of $\ensuremath{\texttt{b}}$.
The scene region $\ensuremath{\texttt{s}}$ is the whole image containing
the head. 

We use a face detector to find the face rectangle $\ensuremath{\texttt{f}}$
inside each test head. We use the open source state of the art method
of \cite{Mathias2014Eccv}, which also provides a rough indication
of the head yaw rotation (frontal, $45\degree$, $90\degree$ side
view). When no detection matches an annotation (e.g. back views),
we regress the face area from the head bounding box. More details
on the performance of this detector are in \S \ref{sec:Head-or-face}.
Five respective image regions are illustrated in Figure \ref{fig:image-regions}.

Please note that regions overlap with each other, and that those pose
agnostic crops may not match the actual regions.

\subsection{\label{sub:Implementation}Fine-tuning and parameters}

Unless specified otherwise AlexNet is fine-tuned using PIPA's person
recognition training set ($\sim\negmedspace30\mbox{k}$ instances,
$\sim\negmedspace1.5\mbox{k}$ identities), cropped at different regions,
with $300\mbox{k}$ mini-batch iterations (batch size $50$). We refer
to the base cue thus obtained as $\ensuremath{\texttt{f}}$, $\ensuremath{\texttt{h}}$,
$\ensuremath{\texttt{u}}$, $\ensuremath{\texttt{b}}$, or $\ensuremath{\texttt{s}}$,
depending on the crop. On the validation set we found fine-tuning
to provide a systematic $\sim\negmedspace10$ percent points (pp)
gain over not fine-tuned AlexNet. Since we use seventh layer of AlexNet,
each cue adds $4\,096$ dimensions to our concatenated feature vector.

We train for each identity linear classifier using SVM regularization
parameter $C=1$. On the validation set the SVM classifier consistently
outperforms by a $\sim\negmedspace10\ \mbox{pp}$ margin the naive
nearest neighbour (NN) classifier. Additional details can be found
in Appendix \S\ref{sec:fine-tuning}.

\subsection{\label{sec:how-informative-body-region}How informative is each image
region ?}

\begin{table}
\begin{centering}
\begin{tabular}{llc}
\multicolumn{2}{l}{Cue} & Accuracy\tabularnewline
\hline 
\hline 
\multicolumn{2}{l}{Chance level} & \hspace{0.5em}0.27\tabularnewline
\hline 
Scene (\S\ref{sec:Scene}) & \texttt{$\texttt{s}$} & 27.06\tabularnewline
Body  & $\mbox{\ensuremath{\texttt{b}}}$ & 80.81\tabularnewline
Upper body & $\texttt{u}$ & 84.76\tabularnewline
Head & $\texttt{h}$ & 83.88\tabularnewline
Face (\S\ref{sec:Head-or-face}) & $\texttt{f}$ & 74.45\tabularnewline
\hline 
Face+head & $\texttt{f}\negthinspace+\negthinspace\texttt{h}$ & 84.80\tabularnewline
Full person & $\texttt{P}=\texttt{f}\negthinspace+\negthinspace\texttt{h}\negthinspace+\negthinspace\texttt{u}\negthinspace+\negthinspace\texttt{b}$\hspace*{-1.5em} & 91.14\tabularnewline
Full image & $\texttt{\ensuremath{\mbox{\ensuremath{\texttt{P}}}_{s}}}=\texttt{P}\negthinspace+\negthinspace\texttt{s}$ & 91.16\tabularnewline
\end{tabular}
\par\end{centering}

\protect\caption{\label{tab:validation-set-regions-accuracy}Validation set accuracy
of different cues. More detailed combinations in Appendix Table \ref{tab:validation-set-regions-accuracy-appendix}.}
\end{table}

Table \ref{tab:validation-set-regions-accuracy} shows the validation
set results of each region individually and in combination. Head and
upper body are the strongest individual cues. We discuss head and
face in \S \ref{sec:Head-or-face}. Upper body is more reliable than
the full body, because we observe that legs are commonly occluded
(or out of frame) and thus become a distractor. Scene is, unsurprisingly,
the weakest individual cue, but it still contains useful information
for person recognition (far above chance level). Importantly, we see
that all cues complement each other (despite having overlapping pixels).

\paragraph{Conclusion}

On the validation set at least, our features and combination strategy
seems quite effective.

\subsection{\label{sec:Scene}Scene ($\textnormal{\texttt{s}}$)}

\begin{table}
\begin{centering}
\begin{tabular}{llc}
 & Method & Accuracy\tabularnewline
\hline 
\hline 
Gist & \texttt{$\texttt{s}_{\texttt{gist}}$} & 21.56\tabularnewline
PlacesNet scores & \texttt{$\texttt{s}_{\texttt{places 205}}$} & 21.44\tabularnewline
raw PlacesNet & \texttt{$\texttt{s}_{0\texttt{ places}}$} & 27.37\tabularnewline
PlacesNet fine-tuned & \texttt{$\texttt{s}_{3\texttt{ places}}$} & 25.62\tabularnewline
raw AlexNet & \texttt{$\texttt{s}_{0}$} & 26.54\tabularnewline
AlexNet fine-tuned & \texttt{$\texttt{s}=\texttt{s}_{3}$} & 27.06\tabularnewline
\end{tabular}
\par\end{centering}

\protect\caption{\label{tab:validation-set-scene}Validation set accuracy of different
feature vectors for the scene region $\textnormal{\texttt{s}}$. See
descriptions in \S \ref{sec:Scene}.}
\end{table}

Other than a fine-tuned AlexNet we considered multiple feature types
to encode the scene information. \texttt{$\texttt{s}_{\texttt{gist}}$}:
using the Gist descriptor \cite{Oliva2001IjcvGist} ($512$ dimensions).
\texttt{$\texttt{s}_{0\texttt{ places}}$}: instead of using AlexNet
pre-trained on ImageNet, we consider an AlexNet (PlacesNet) pre-trained
on $205$ scene categories of the ``Places Database'' \cite{Zhou2014NipsPlaces}
($\sim\negmedspace2.5$ million images). \texttt{$\texttt{s}_{\texttt{places 205}}$}:
Instead of the $4\,096$ dimensions PlacesNet feature vector, we also
consider using the score vector for each scene category ($205$ dimensions).
$\texttt{s}_{0}$,$\texttt{s}_{3}$: we consider using AlexNet in
the same way as for body or head (with zero or $300\mbox{k}$ iterations
of fine-tuning on the PIPA person recognition training set). \texttt{$\texttt{s}_{3\texttt{ places}}$: $\texttt{s}_{0\texttt{ places}}$
}fine-tuned for person recognition.

\paragraph{Results}

Table \ref{tab:validation-set-scene} compares the different alternatives
on the validation set. The Gist descriptor \texttt{$\texttt{s}_{\texttt{gist}}$}
performs only slightly below the convnet options ($4\,608$ dimensional
version of Gist gives worse results). Using the raw (and longer)
feature vector of \texttt{$\texttt{s}_{0\texttt{ places}}$} is better
than the class scores of \texttt{$\texttt{s}_{\texttt{places 205}}$}.
Interestingly, in this context pre-training for places classification
is better than pre-training for objects classification (\texttt{$\texttt{s}_{0\texttt{ places}}$}
versus \texttt{$\texttt{s}_{0}$}). After fine-tuning $\texttt{s}_{3}$
reaches a similar performance as \texttt{$\texttt{s}_{0\texttt{ places}}$}.\\
Experiments trying different combinations indicate that there is little
complementarity between these features. Since there is not a large
difference between \texttt{$\texttt{s}_{0\texttt{ places}}$} and
\texttt{$\texttt{s}_{3}$}, for sake of simplicity we use \texttt{$\texttt{s}_{3}$}
as our scene cue \texttt{$\texttt{s}$} in all other experiments.

\paragraph{Conclusion}

Scene by itself, albeit weak, can obtain results far above chance
level. After fine-tuning, scene recognition as pre-training surrogate
task \cite{Zhou2014NipsPlaces} does not provide a clear gain over
(ImageNet) object recognition.

\subsection{\label{sec:Head-or-face}Head ($\textnormal{\texttt{h}}$) or face
($\textnormal{\ensuremath{\texttt{f}}}$) ?}

A large portion of work on face recognition focuses on the face region
specifically. In the context of photo albums, we aim to quantify how
much information is available in the head versus the face region.

The face region $\ensuremath{\texttt{f}}$ is defined by a state of
the art face detector \cite{Mathias2014Eccv} (see \S \ref{sub:Body-regions}).
Since no face annotations are available on PIPA, we validate the face
detection location by learning a linear regressor from $\ensuremath{\texttt{f}}$
to $\ensuremath{\texttt{h}}$ (per DPM component). When using these
heads estimates ($\sim\negmedspace75\%$ of heads replaced) instead
of the ground truth head ($\ensuremath{\texttt{h}}$ in Table \ref{tab:validation-set-regions-accuracy}),
results drop only $0.45\%$ thus indirectly validating that faces
are well localized.

\paragraph{Results}

When using the face region, there is a large gap of $\sim\negmedspace10$
percent points performance between $\texttt{f}$ and $\texttt{h}$
in Table \ref{tab:validation-set-regions-accuracy} highlighting the
importance of including the head region around the face in the descriptor.

When evaluating only on the frontal faces of validation set (as indicated
by the detector) $\texttt{f}$ reaches $81\%$ accuracy and $70\%$
for non-frontal faces. The performance drop between frontal versus
handling profile and back views is less dramatic than one could have
suspected.

In comparison, on frontal faces in test set, \texttt{DeepFace} reaches
$\sim\negmedspace90\%$ \cite{Zhang2015CvprPiper}, and returns the
chance level ($0.17\%$) otherwise. The test set contains about $50\%$
of non-frontal faces. On test set $\texttt{f}$ obtains $74\%$ and
$57\%$ for frontal and non-frontal faces, respectively ($18\ \mbox{pp}$
drop), while $\texttt{h}$ obtains $82\%$ and $70\%$, respectively
($12\ \mbox{pp}$ drop).

\paragraph{Conclusion}

Using $\texttt{h}$ is more effective than $\texttt{f}$, both due
to improved recognition for frontal faces and robustness to head rotation.
That being said, $\texttt{f}$ results show fair performance to recognise
non-frontal faces. As with other body cues, there is complementarity
between $\texttt{h}$ and $\texttt{f}$ and we thus suggest to use
them together.

\begin{table}
\begin{centering}
\begin{tabular}{llc}
 & Method & Accuracy\tabularnewline
\hline 
\hline 
More data (\S\ref{sec:Additional-training-data}) & $\texttt{h}$ & 83.88\tabularnewline
{\footnotesize{}(head region)} & $\texttt{h}+\mbox{\ensuremath{\texttt{h}}}_{\texttt{cacd}}$ & 84.88\tabularnewline
 & $\texttt{h}+\mbox{\ensuremath{\texttt{h}}}_{\texttt{casia}}$ & 86.08\tabularnewline
 & $\texttt{h}+\mbox{\ensuremath{\texttt{h}}}_{\texttt{casia}}+\mbox{\ensuremath{\texttt{h}}}_{\texttt{cacd}}$\hspace*{-1.5em} & 86.26\tabularnewline
\hline 
Attributes (\S\ref{sec:Attributes})  & $\mbox{\ensuremath{\texttt{h}}}_{\texttt{pipa11m}}$  & 74.63\tabularnewline
{\footnotesize{}(head region)} & $\mbox{\ensuremath{\texttt{h}}}_{\texttt{pipa11}}$  & 81.74\tabularnewline
 & $\texttt{h}+\mbox{\ensuremath{\texttt{h}}}_{\texttt{pipa11}}$ & 85.00\tabularnewline
\arrayrulecolor{gray}\hline\arrayrulecolor{black}{\footnotesize{}(upper
body region)} & $\mbox{\ensuremath{\texttt{u}}}_{\texttt{peta5}}$ & 77.50\tabularnewline
 & $\texttt{u}+\mbox{\ensuremath{\texttt{u}}}_{\texttt{peta5}}$ & 85.18\tabularnewline
\arrayrulecolor{gray}\hline\arrayrulecolor{black}{\footnotesize{}(head+upper
body)} & $\mbox{\ensuremath{\texttt{A}}}=\mbox{\ensuremath{\texttt{h}}}_{\texttt{pipa11}}+\mbox{\ensuremath{\texttt{u}}}_{\texttt{peta5}}$\hspace*{-1.5em} & 86.17\tabularnewline
 & $\texttt{h}+\texttt{u}$ & 85.77\tabularnewline
 & $\texttt{h}+\texttt{u}+\ensuremath{\texttt{A}}$ & 90.12\tabularnewline
\end{tabular}
\par\end{centering}

\protect\caption{\label{tab:validation-set-extended-data-accuracy}Validation set accuracy
of different cues based on extended data. See \S\ref{sec:Additional-training-data}
and \S\ref{sec:Attributes} for details.}
\end{table}

\subsection{\label{sec:Additional-training-data}Additional training data ($\textnormal{\ensuremath{\mbox{\ensuremath{\texttt{h}}}_{\texttt{cacd}}},\,\ensuremath{\mbox{\ensuremath{\texttt{h}}}_{\texttt{casia}}}}$)}

It is well known that deep learning architectures benefit from additional
data. \texttt{PIPER}'s DeepFace is trained over $4.4\cdot10^{6}$
faces of $4\cdot10^{3}$ persons (the private SFC dataset \cite{Taigman2014CvprDeepFace}).
In comparison our cues are trained over ImageNet and PIPA's $29\cdot10^{3}$
faces over $1.4\cdot10^{3}$ persons. To measure the effect of training
on larger data we consider fine-tuning using two open face recognition
datasets: CASIA-WebFace (CASIA) \cite{Yi2014ArxivLearningFace} and
the ``Cross-Age Reference Coding Dataset'' (CACD) \cite{Chen2014Eccv}.

CASIA contains $0.5\cdot10^{6}$ images of $10.5\cdot10^{3}$ persons
(mainly actors and public figures), and is (to the best of our knowledge)
the largest open dataset for face recognition. When fine-tuning AlexNet
over these identities (using the head area $\mbox{\ensuremath{\texttt{h}}}$),
we obtain the $\mbox{\ensuremath{\texttt{h}}}_{\texttt{casia}}$ cue. 

CACD contains $160\cdot10^{3}$ faces of $2\cdot10^{3}$ persons 
with varying ages. Although smaller than CASIA, CACD features greater
number of face examples per subject ($\sim\negmedspace2\times)$.
The $\mbox{\ensuremath{\texttt{h}}}_{\texttt{cacd}}$ cue is built
via the same procedure as $\mbox{\ensuremath{\texttt{h}}}_{\texttt{casia}}$.

\paragraph{Results}

The improvement of $\texttt{h}+\mbox{\ensuremath{\texttt{h}}}_{\texttt{cacd}}$
and $\texttt{h}+\mbox{\ensuremath{\texttt{h}}}_{\texttt{casia}}$
over $\texttt{h}$ show that cues from outside training data are complementary
to $\texttt{h}$ (see top part of Table \ref{tab:validation-set-extended-data-accuracy}).
$\mbox{\ensuremath{\texttt{h}}}_{\texttt{cacd}}$ and $\mbox{\ensuremath{\texttt{h}}}_{\texttt{casia}}$
on their own are about $\sim\negmedspace5\ \mbox{pp}$ worse than
$\texttt{h}$. $\mbox{\ensuremath{\texttt{h}}}_{\texttt{cacd}}$
and $\mbox{\ensuremath{\texttt{h}}}_{\texttt{casia}}$ exhibit slight
complementarity.

\paragraph{Conclusion}

Adding more data, even from different type of photos, is an effective
means to improve the performance.

\subsection{\label{sec:Attributes}Attributes ($\textnormal{\texttt{\ensuremath{\mbox{\ensuremath{\texttt{h}}}_{\texttt{pipa11}}},\,\ensuremath{\mbox{\ensuremath{\texttt{u}}}_{\texttt{peta5}}}}}$)}

Albeit overall appearance might change day to day, one could expect
that long term attributes provide means for recognition. We thus explore
building feature vectors by fine-tuning AlexNet not on person recognition
(like for all other cues), but rather for attributes classification
as a surrogate task. We consider two sets of annotations.

We have annotated the PIPA train and validation sets ($1409+366$
identities) with five long term attributes: age, gender, glasses,
hair colour, and hair length ($11$ binary bits total; see Appendix
\S\ref{sec:Attributes-supp} for details). We use the $\texttt{h}$
crops to build $\mbox{\ensuremath{\texttt{h}}}_{\texttt{pipa11}}$,
as the attributes are head centric.

We also consider using the ``PETA pedestrian attribute dataset''
\cite{Deng2014AcmPeta}, which features $105$ attributes annotations
for $19\cdot10^{3}$ full-body pedestrian images. Out of $105$ we
chose the five binary attributes that are long term and are well represented
in PETA: gender, age (young adult, adult), black hair, and short
hair (details in Appendix \S\ref{sec:Attributes-supp}). Since upper-body
$\texttt{u}$ is less noisy than the full-body $\mbox{\ensuremath{\texttt{b}}}$
(see Table \ref{tab:validation-set-regions-accuracy}), upper body
crops of PETA are used to fine-tune AlexNet. The $\mbox{\ensuremath{\texttt{u}}}_{\texttt{peta5}}$
cue is thus built.

\paragraph{Results}

For attribute fine-tuning we consider two approaches: training a single
network for all attributes (multi-label classification with sigmoid
cross entropy loss), or tuning one separate network per attribute
(softmax loss) and then concatenating their feature vectors. The results
on validation data indicate the second choice ($\mbox{\ensuremath{\texttt{h}}}_{\texttt{pipa11}}$)
performs better than the first  ($\mbox{\ensuremath{\texttt{h}}}_{\texttt{pipa11m}}$).

Table \ref{tab:validation-set-extended-data-accuracy} (bottom) shows
that attribute classification as surrogate task does help person recognition.
Both PIPA ($\mbox{\ensuremath{\texttt{h}}}_{\texttt{pipa11}}$) and
PETA ($\mbox{\ensuremath{\texttt{u}}}_{\texttt{peta5}}$) annotations
behave similarly ($\sim\negmedspace1\ \mbox{pp}$ gain over $\texttt{h}$
and $\texttt{u}$), and show good complementary among themselves ($\sim\negmedspace5\ \mbox{pp}$
gain over $\texttt{h}\negmedspace+\negmedspace\texttt{u}$). Amongst
the attributes considered, gender contributes the most to improve
recognition accuracy (for both attributes datasets).

\paragraph{Conclusion}

Adding attributes classification as a surrogate task improves performance.

\section{\label{sec:Test-set-results}Test set results}

\begin{table}
\begin{centering}
\begin{tabular}{llc}
 & Method & Accuracy\tabularnewline
\hline 
\hline 
\arrayrulecolor{gray} & Chance level & \hspace{0.5em}0.17\tabularnewline
\hline 
Body & \texttt{GlobalModel} \texttt{\cite{Zhang2015CvprPiper}} & 67.60\tabularnewline
 & $\mbox{\ensuremath{\texttt{b}}}$  & 69.63\tabularnewline
\hline 
Head & \texttt{DeepFace \cite{Zhang2015CvprPiper}} & 46.66\tabularnewline
 & $\mbox{\ensuremath{\texttt{h}}}$ & 76.42\tabularnewline
{\small{}Extended data} & $\texttt{h}+\mbox{\ensuremath{\texttt{h}}}_{\texttt{casia}}+\mbox{\ensuremath{\texttt{h}}}_{\texttt{cacd}}$\hspace*{-1.5em} & 79.63\tabularnewline
\hline 
\arrayrulecolor{black}\hline\arrayrulecolor{gray} & \texttt{PIPER \cite{Zhang2015CvprPiper}} & 83.05\tabularnewline
{\small{}Head+Body} & $\texttt{h}\negthinspace+\negthinspace\texttt{b}$ & 83.36\tabularnewline
{\small{}Full person} & $\texttt{P}=\texttt{f}\negthinspace+\negthinspace\texttt{h}\negthinspace+\negthinspace\texttt{u}\negthinspace+\negthinspace\texttt{b}$\hspace*{-1.5em} & 85.33\tabularnewline
{\small{}Full image} & $\texttt{\ensuremath{\mbox{\ensuremath{\texttt{P}}}_{s}}}=\texttt{P}\negthinspace+\negthinspace\texttt{s}$ & 85.71\tabularnewline
{\small{}Extended data} & $\texttt{naeil}=\texttt{\ensuremath{\mbox{\ensuremath{\texttt{P}}}_{s}}}\negthinspace+\negthinspace\texttt{E}$ & \textbf{86.78}\tabularnewline
\hline 
{\small{}Combining } & $\texttt{PIPER}$\cite{Zhang2015CvprPiper}$\negthinspace+\negthinspace\texttt{P}$ & 87.67\tabularnewline
{\small{}with }\texttt{\small{}\cite{Zhang2015CvprPiper}} & $\texttt{PIPER}$\cite{Zhang2015CvprPiper}$\negthinspace+\negthinspace\texttt{naeil}$ & 88.37\tabularnewline
\end{tabular}
\par\end{centering}

\protect\caption{\label{tab:test-set-accuracy}Test set accuracy of different cues
and their combinations under the original PIPA evaluation protocol.\protect \\
Extended data $\texttt{E}=\mbox{\ensuremath{\texttt{h}}}_{\texttt{casia}}\negthinspace+\negthinspace\mbox{\ensuremath{\texttt{h}}}_{\texttt{cacd}}\negthinspace+\negthinspace\texttt{\ensuremath{\mbox{\ensuremath{\texttt{h}}}_{\texttt{pipa11}}}+\ensuremath{\mbox{\ensuremath{\texttt{u}}}_{\texttt{peta5}}}}$.}
\end{table}

All experiments in this paper are limited to a person recognition scenario where head boxes are provided by human annotations, and all test faces belong to a known finite set.
Table \ref{tab:test-set-accuracy} reports the performance on the
test set of the different cues described in previous sections. We
study their complementarity to each other, and compare them against
the \texttt{PIPER} components \cite{Zhang2015CvprPiper}. A more detailed
table and the corresponding validation set results are included in
Appendix Table \ref{tab:splits-accuracy}.

We also report computational times for some pipelines in our method. The feature training takes 2-3 days on a single GPU machine. The SVM training takes 42.20s for $\texttt{h}$ (4096 dim) and 1303.30s for $\texttt{naeil}$ ($4096\times 17$ dim) on the Original split (581 classes, 6443 samples). Note that this corresponds to a realistic user scenario in a photo sharing service where $\sim 500$ identities are known to the user and the average number of photos per identity is $\sim 10$.

Compared to PIPER, our framework is computationally efficient in two aspects. First, our system does not need to learn to assign weights for different cues. Second, the PIPER feature has roughly $4096\times 108$ dimensions, requiring far more memory and training time than our final system ($4096\times 17$ dim).

\paragraph{Body $\textnormal{\texttt{b}}$}

Considered alone, our body cue $\mbox{\ensuremath{\texttt{b}}}$
is a reimplementation of \texttt{PIPER}'s \texttt{GlobalModel} \cite{Zhang2015CvprPiper}.
As expected, we obtain a similar accuracy.

\paragraph{Head $\textnormal{\texttt{h}}$}

On the other hand, our head cue $\textnormal{\texttt{h}}$ is more
effective than the corresponding \texttt{PIPER}'s \texttt{DeepFace}.
As discussed in \S \ref{sec:Head-or-face}, we have observed that:
a) for this task the head region is more informative than the face
(focusing on the face region is detrimental); b) our approach is much
more robust for non-frontal faces ($\sim\negmedspace50\%$ of test
cases), where $\textnormal{\texttt{h}}$ reaches $70\%$ accuracy,
while \texttt{DeepFace} becomes uninformative in this case. When extending
the training data our head performance further improves (see also
the discussion in \S \ref{sub:Analysis-of-failure}).

\paragraph{Head+body $\textnormal{\texttt{\texttt{h}\negthinspace+\negthinspace\texttt{b}}}$}

Our minimal system matching \texttt{PIPER}'s performance is $\texttt{h}\negthinspace+\negthinspace\texttt{b}$,
with accuracy $83.36\%$. Note that the feature vector of $\texttt{h}\negthinspace+\negthinspace\texttt{b}$
is about $50$ times smaller than \texttt{PIPER}'s. 

In principle \texttt{PIPER} captures the head region via one of its
poselets. Thus, $\texttt{h}\negthinspace+\negthinspace\texttt{b}$
extracts cues from a subset of \texttt{PIPER}'s ``\texttt{GlobalModel+Poselets}''
\cite{Zhang2015CvprPiper}, which only reaches $78.79\%$ .

\paragraph{Full person $\textnormal{\texttt{P}}$}

Similar to the validation set results, having more cues further improves
results. $\texttt{P}=\texttt{f}\negthinspace+\negthinspace\texttt{h}\negthinspace+\negthinspace\texttt{u}\negthinspace+\negthinspace\texttt{b}$
obtains a clear margin over \texttt{PIPER}, yet is a simpler system
(neither specialised face recognition nor pose estimation used) built
with less training data (only PIPA for fine-tuning, ImageNet for pre-training,
and the face detector training set).

\paragraph{$\texttt{naeil}$}

Adding scene $\textnormal{\texttt{s}}$ (\S\ref{sec:Scene}) and
extended data $\textnormal{\texttt{E}}$ (\S\ref{sec:Additional-training-data}
\& \S\ref{sec:Attributes}) contributes to the last $1$ percent
point. We name our final method $\texttt{naeil}$\footnote{``naeil'', \includegraphics[height=0.8em]{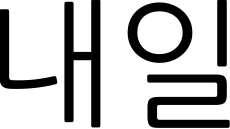},
means ``tomorrow'' and sounds like ``nail''.}. Its feature vector is $6$ times smaller than \texttt{PIPER}'s,
and it provides the best known results on the PIPA dataset.

Figures \ref{fig:teaser} and \ref{fig:success-O-split} show some
example results of our system. \S \ref{sub:Analysis-of-failure}
analyses the remaining hard test cases.

\subsection{\label{sub:Combining-methods}Complementarity between \texttt{PIPER}
and \texttt{naeil}}

Since $\texttt{PIPER}$ uses different training data than $\texttt{naeil}$
we can expect some complementarity between the two methods. For experiments, we use the $\texttt{PIPER}$ scores provided by the authors of \cite{Zhang2015CvprPiper}. Note, however, that the $\texttt{PIPER}$ features are unavailable.
By averaging the output scores of the two methods ($\texttt{PIPER}\negthinspace+\negthinspace\texttt{naeil}$)
gain $\sim\negmedspace1.5\ \mbox{percent points}$, reaching $88.37\%$.
Using a more sophisticated strategy might provide more gain, but we
already see that $\texttt{naeil}$ covers most of the performance
from $\texttt{PIPER}$.

\subsection{\label{sub:Towards-open-world}Towards an open world setting}

All experiments in this paper are limited to a person recognition
scenario where head boxes are provided by human annotations, and all
test faces belong to a known finite set. Not providing ground truth
heads at test time is an arguably more realistic and challenging scenario
in which both person detection and recognition need to be solved jointly.

Using a face detector (\S\ref{sec:Head-or-face}) as our person detector
over the test set, we reach $\sim\negmedspace78\%$ recall at (average)
ten detections per image ($\sim\negmedspace70\%$ at 3 $\nicefrac{\mbox{detections}}{\mbox{image}}$).
If we use $\texttt{naeil}$ to label these faces, we reach $\sim\negmedspace65\%$
recall on the $\mbox{test}_{\nicefrac{0}{1}}$ identities ($\sim\negmedspace62\%$
at 3 $\nicefrac{\mbox{detections}}{\mbox{image}}$). 

The performance drops, but less dramatically than what one might expect.
It remains as future work to implement a detailed evaluation in the
open world setting.

\subsection{\label{sub:face-rgb-baseline}A naive baseline}

\begin{figure*}
\begin{centering}
\includegraphics[bb=0bp 0bp 1820bp 1046bp,width=2\columnwidth]{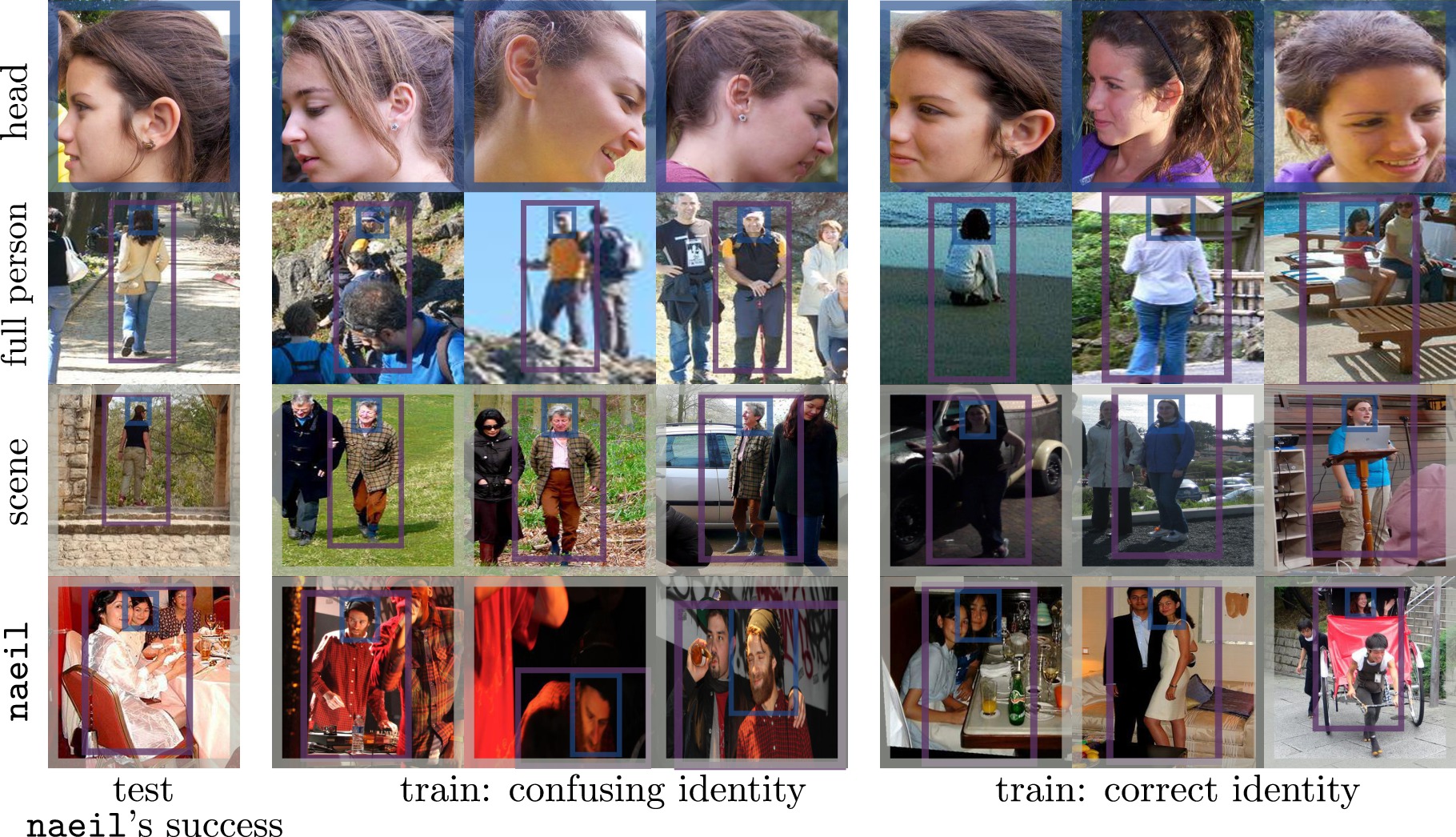}
\par\end{centering}

\begin{raggedright}
\medskip{}

\par\end{raggedright}

\protect\caption{\label{fig:success-O-split} Examples of success cases on the Original
split. First column shows the test instances that our systems correctly
predict. Columns 5-7 correspond to train instances of the correct
identity. Columns 2-4 are the training examples of the identity that
\texttt{PIPER} \cite{Zhang2015CvprPiper} wrongly predicts. From top
to bottom, the shown test instances are: (1) success case of $\mbox{\ensuremath{\texttt{f}}}\negthinspace+\negthinspace\mbox{\ensuremath{\texttt{h}}}$
and failure case of \texttt{PIPER}; (2) success case of $\mbox{\ensuremath{\texttt{p}}}\negthinspace=\negthinspace\mbox{\ensuremath{\texttt{f}}}\negthinspace+\negthinspace\mbox{\ensuremath{\texttt{h}}}\negthinspace+\negthinspace\mbox{\ensuremath{\texttt{u}}}\negthinspace+\negthinspace\mbox{\ensuremath{\texttt{b}}}$
and failure case of \texttt{PIPER} and $\mbox{\ensuremath{\texttt{f}}}\negthinspace+\negthinspace\mbox{\ensuremath{\texttt{h}}}$;
(3) success case of $\mbox{\ensuremath{\texttt{p}}}\negthinspace+\negthinspace\mbox{\ensuremath{\texttt{s}}}$
and failure case of \texttt{PIPER and} $\mbox{\ensuremath{\texttt{p}}}$;
and (4) success case of \texttt{naeil},and failure case of \texttt{PIPER
and} $\mbox{\ensuremath{\texttt{p}}}\negthinspace+\negthinspace\mbox{\ensuremath{\texttt{s}}}$.}
\vspace{-1em}
\end{figure*}

Given the inherent difficulty of the PIPA person recognition task
(see Figure \ref{fig:success-O-split}) reaching a $\sim\negmedspace85\%$
accuracy seems suspiciously high. Thus, we investigate the issue using
a crude baseline $\mbox{\ensuremath{\texttt{h}}}_{\texttt{rgb}}$
that takes the raw RGB pixel values of the head area as features (after
downsizing to $40\negmedspace\times\negmedspace40\ \mbox{pixels}$
and blurring), and uses a nearest neighbour classifier. By design
$\mbox{\ensuremath{\texttt{h}}}_{\texttt{rgb}}$ is only able to recognize
near identical heads across the $\mbox{test}_{\nicefrac{0}{1}}$split,
yet it reaches a surprisingly high $33.77\%$ ($49.46\%$) accuracy
on the test set (validation set).

\paragraph{Conclusion}

About $\nicefrac{1}{3}$ of the original PIPA test splits is easy
to solve. This motivates us to explore more realistic splits and protocols.
In the next section we discuss the issue and propose solutions via
new test splits.

\begin{figure}
\begin{centering}
\hspace*{\fill}\includegraphics[width=1\columnwidth]{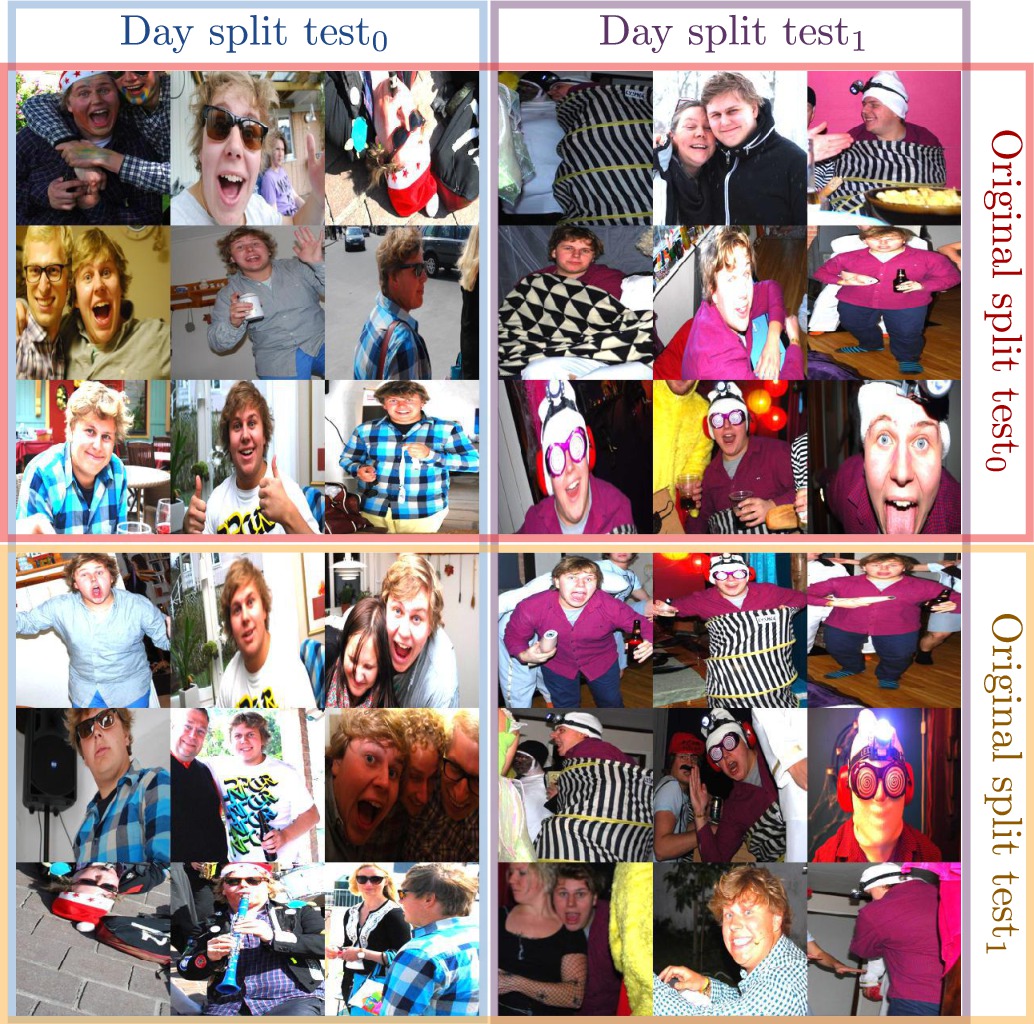}\hspace*{\fill}\vspace{0.5em}

\par\end{centering}

\protect\caption{\label{fig:splits-visualisation}Visualisation of Original and Day
splits for one identity. Greater appearance changes are observed across
the Day split.}
\end{figure}

\section{\label{sec:failures-analysis}Analysis of person recognition challenges}

This section provides a detailed analysis of the obtained results
and shares insights on addressing future challenges. 

As we have seen in \S\ref{sub:face-rgb-baseline}, the current setup
includes many easy examples, limiting us from exploring more difficult
dimensions of the problem. Accordingly, we propose three new $\mbox{test}_{0}$/
$\mbox{test}_{1}$ splits of PIPA in \S\ref{sub:PIPA-splits}. Based
on the new splits, we analyse the robustness of different cues across
appearance changes (\S\ref{sub:Importance-of-features}). We then
discuss the effect of the amount of person specific training data
(\S\ref{sub:Importance-of-training}), and provide a failure mode
analysis in \S  \ref{sub:Analysis-of-failure}.

\begin{figure*}
\centering{}\hspace*{\fill}%
\begin{minipage}[t]{0.3\textwidth}%
\begin{center}
\includegraphics[height=1\columnwidth]{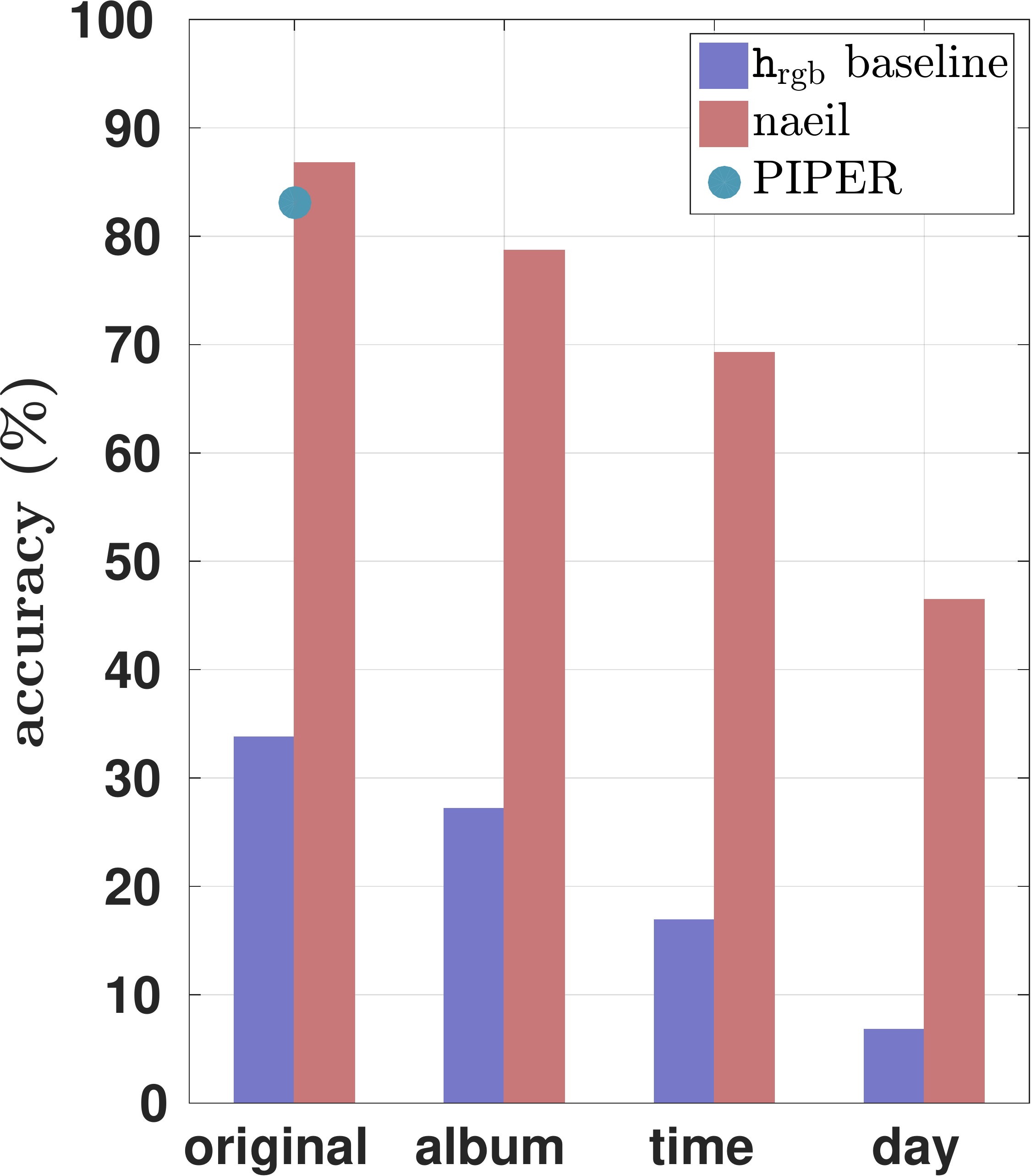}
\par\end{center}

\begin{center}
\protect\caption{\label{fig:splits-accuracy}Recognition accuracy across different
experimental setups on test set.}

\par\end{center}%
\end{minipage}\hspace*{\fill}%
\begin{minipage}[t]{0.3\textwidth}%
\begin{center}
\vspace{-14.8em}
\includegraphics[width=1\textwidth,height=1\textwidth]{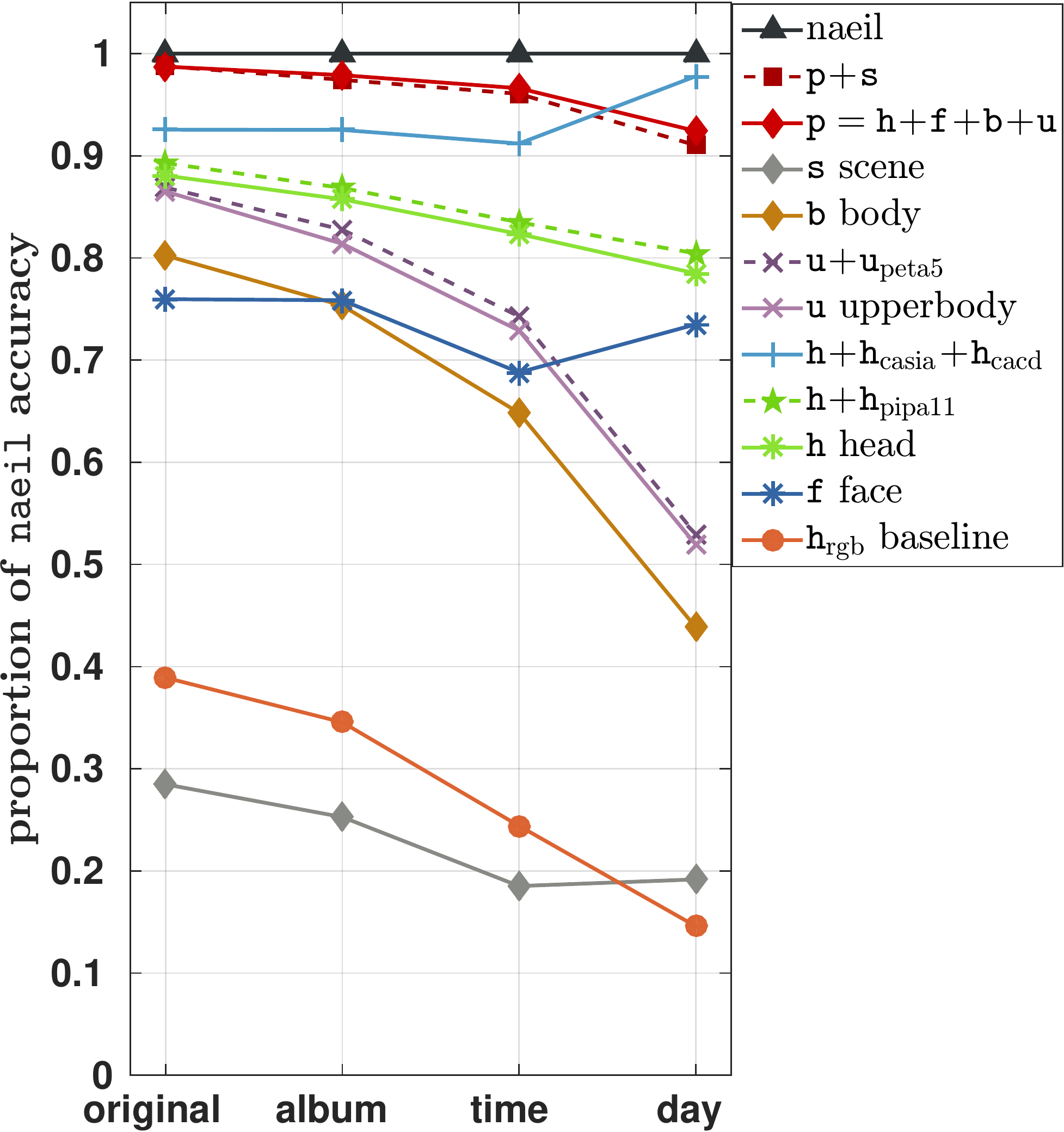}
\par\end{center}

\begin{center}
\vspace{-0.2em}
\protect\caption{\label{fig:splits-accuracy-relative}Test set accuracy of cues in
different settings, relative to \texttt{naeil}.}

\par\end{center}%
\end{minipage}\hspace*{\fill}%
\begin{minipage}[t]{0.3\textwidth}%
\begin{center}
\includegraphics[height=1\columnwidth]{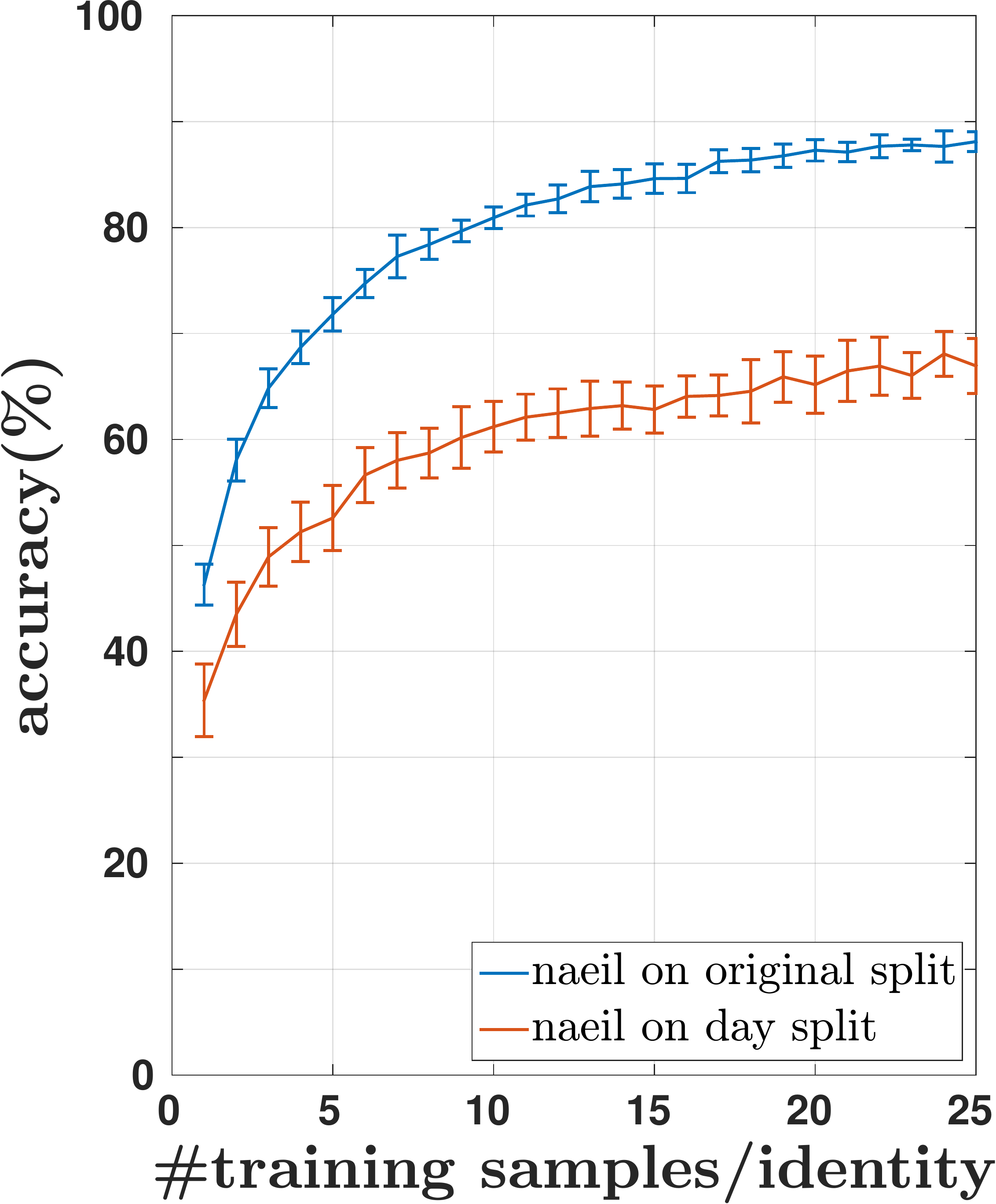}\hspace{0.9em}
\par\end{center}

\begin{center}
\protect\caption{\label{fig:numtrain-accuracy}Recognition accuracy at different sizes
of training examples.}

\par\end{center}%
\end{minipage}\hspace*{\fill}\vspace{-1.5em}
\end{figure*}

\subsection{\label{sub:PIPA-splits}New PIPA splits with varying difficulty and
challenges}

We have seen a strong performance of our main system \texttt{naeil
}($86.78\%$ on test set, Table \ref{tab:test-set-accuracy}) and
the baseline $\mbox{\ensuremath{\texttt{h}}}_{\texttt{rgb}}$ (33.77\%
on test set, \S\ref{sub:face-rgb-baseline}) despite the challenging
task of person recognition in photo albums. This motivates us to investigate
more difficult and realistic setups.

\paragraph{Limitations of original setup}

The main limitation of the original PIPA protocol is that the $\mbox{test}_{0}$/
$\mbox{test}_{1}$ splits are even-odd instances from a sample list
that largely preserves the photo orders in albums. When photos are
taken in a short period of time, adjacent photos can be nearly identical.
However, a main challenge in person recognition is to generalise across
long-term appearance changes of a person; we thus introduce a range
of new splits on PIPA in the order of increasing difficulty:

\paragraph{Original split $\mathcal{O}$: }

We keep the original split in our study for comparison. The split
is on the odd vs even basis.

\paragraph{Album split $\mathcal{A}$:}

All samples are organised by albums. This split assigns for each person
identity samples from separate albums, while keeping the number of
samples equal for the splits. Since it is not always possible to satisfy
both conditions, a few albums are shared between the splits. In this
setup, training and test samples are split across different events
and occasions.

\paragraph{Time split $\mathcal{T}$:}

This split investigates the temporal dimension of the photos. For
each person identity, we sort all photos by their ``photo-taken-date''
metadata. We split them into newest versus oldest images. The instances
without time metadata are distributed evenly. This split emphasises
the temporal distance between training and test.

\paragraph{Day split $\mathcal{D}$: }

$\mathcal{T}$ does not always make a time gap: many people appear
only on one event, and the time metadata are often missing. We thus
make the split manually according to the two conditions: either a
firm evidence of date change such as \{change of season, continent,
event, co-occurring people\} between the splits, or visible changes
in \{hairstyle, make-up, head or body wear\}. These rules enforce
``appearance changes''. For each identity, we randomly discard instances
from the larger test set until sizes match. If there are less than
$5$ instances in the split, we discard the identity altogether (Original
split applies the same criterion). After pruning, $199$ identities
(out of $581$) were left, with about $20$ training samples per identity
(similar range as all other splits).

\paragraph{Results}

Figure \ref{fig:splits-accuracy} provides an overview of how the
raw colour baseline $\mbox{\ensuremath{\texttt{h}}}_{\texttt{rgb}}$
and our\texttt{ naeil} approach perform across different splits. We
observe that the unreasonably good performance by the $\mbox{\ensuremath{\texttt{h}}}_{\texttt{rgb}}$
baseline consistently degrades from the Original over Album and Time
to the Day splits, indicating the increasing amount of non-trivial
recognition tasks. Compared to the $1/5$ drop by the $\mbox{\ensuremath{\texttt{h}}}_{\texttt{rgb}}$
baseline (33.77\% to 6.78\%), \texttt{naeil}'s performance is less
impaired (86.78\% to 46.48\%), indicating \texttt{naeil}'s ability
to address more realistic scenarios characterised by changes in appearance,
location and time.

\subsection{Importance of features \label{sub:Importance-of-features}}

To gain a deeper understanding of relative importance of different
cues and their robustness across splits, we consider Figure \ref{fig:splits-accuracy-relative}
which shows the results normalised by the performance of \texttt{naeil}
($100\%$). This allows us to analyse which features maintain, loose
or gain discriminative power when moving from the easier to the more
challenging settings. 

We observe the strongest drops in relative performance for body and
upper body features, due to the loss of discriminability of global
features (e.g. clothing). We see consistent gains for using surrogate
training tasks such as attributes ($\mbox{\ensuremath{\texttt{h}}}_{\texttt{pipa11}}$,
$\mbox{\ensuremath{\texttt{u}}}_{\texttt{peta5}}$) and, more prominently,
external data for head features ($\mbox{\ensuremath{\texttt{h}}}_{\texttt{casia}}$,
$\mbox{\ensuremath{\texttt{h}}}_{\texttt{cacd}}$). External data
for head features particularly pays off for the most difficult day
split.

\paragraph{Conclusion}

The usage of significantly larger databases improves the robustness
of our features, enabling recognition in the most challenging scenarios.

\subsection{Importance of training data \label{sub:Importance-of-training}}

We also investigate how much collecting more data from each person
identity can help to improve performance. In Figure \ref{fig:numtrain-accuracy}
we compare the Original to the Day split and show performance for
different sizes of training samples. While on the original split already
after 10 training examples 80\% performance is reached, the performance
on the Day split sees a relatively slow improvement and stays below
70\% with 25 samples (lagging 20\% behind the Original split).

\paragraph{Conclusion}

From Figure \ref{fig:numtrain-accuracy} we see that only increasing
the training data will struggle to solve the harder Day split. Better
features and better methods are required.

\subsection{Analysis of remaining failure modes \label{sub:Analysis-of-failure}}

In Appendix \S\ref{sec:failure-modes} we provide detailed statistics
to study failure modes in the Original and Day splits. We discuss
here the main findings.

As expected, non-frontal faces are common failure cases for \texttt{naeil}'s
in both Original and Day splits ($\sim\negmedspace50\%$). For frontal
faces, we observe in the Day split a larger proportion of failures
than in the Original split. Even more, the majority of failures correspond
to large heads ($\mbox{height}>100\ \mbox{pixels}$), where good features
can be extracted. To handle better more realistic scenarios it is
thus important to improve the recognition of frontal faces across
diverse settings and long time-spans.

Another interesting aspect is that while \texttt{naeil} on the Original
split has only one identity (out of 581) which is never correctly
predicted, on the Day split the proportion of never correct identities
jumps to $20\%$.  This suggests that there are inherently difficult
identities that our simplistic\texttt{} system currently cannot handle.

\section{\label{sec:Conclusion}Conclusion}

We analysed the problem of person recognition in photo albums where
people appear with various viewpoints, poses and occlusions. There
are four major conclusions from our studies. First, head region, even
when face is not visible, is a strong cue for person recognition,
better than the face region itself (\S\ref{sec:Head-or-face}). Second,
different cues, although from overlapping regions, are complementary
(\S\ref{sec:Test-set-results}). Third, feature learning with massive
database of faces improves robustness across time and appearance (\S\ref{sub:Importance-of-features}).
Fourth, simply increasing the number of training examples per person
does not automatically solve the problem, and better recognition systems
must be devised (\S\ref{sub:Importance-of-training}). 

One possible research direction is collecting a large database of
personal photo albums on which better features can be trained. One
can also exploit album context, which is a rich source of identity
information \cite{Gallagher2007Cvpr,Shi2013Iccv,Stone2008Cvprw};
however, it was not used in this work for fair comparison.

Our experimental data, including the new splits, trained models, $\mathtt{naeil}$
results, and attribute annotations are published at \textcolor{pink}{http://goo.gl/DKuhlY}.

\begin{otherlanguage}{english}

\bibliographystyle{ieee}
\bibliography{2015_iccv_person_recognition}

\begin{thebibliography}{10}\itemsep=-1pt

\bibitem{Bak2014WacvBrownian}
S.~Bak, R.~Kumar, and F.~Bremond.
\newblock Brownian descriptor: A rich meta-feature for appearance matching.
\newblock In {\em WACV}, 2014.

\bibitem{Bedagkar2014IvcPersonReIdSurvey}
A.~Bedagkar-Gala and S.~K. Shah.
\newblock A survey of approaches and trends in person re-identification.
\newblock {\em IVC}, 2014.

\bibitem{Benfold2009BmvcTownCentre}
B.~Benfold and I.~Reid.
\newblock Guiding visual surveillance by tracking human attention.
\newblock In {\em BMVC}, 2009.

\bibitem{Bourdev2009IccvPoselets}
L.~Bourdev and J.~Malik.
\newblock Poselets: Body part detectors trained using 3d human pose
  annotations.
\newblock In {\em ICCV}, 2009.

\bibitem{Cao2013IccvTransferLearning}
X.~Cao, D.~Wipf, F.~Wen, and G.~Duan.
\newblock A practical transfer learning algorithm for face verification.
\newblock In {\em ICCV}, 2013.

\bibitem{Chen2014Eccv}
B.-C. Chen, C.-S. Chen, and W.~H. Hsu.
\newblock Cross-age reference coding for age-invariant face recognition and
  retrieval.
\newblock In {\em ECCV}, 2014.

\bibitem{Chen2013CvprBlessing}
D.~Chen, X.~Cao, F.~Wen, and J.~Sun.
\newblock Blessing of dimensionality: High-dimensional feature and its
  efficient compression for face verification.
\newblock In {\em CVPR}, 2013.

\bibitem{Cheng2011Bmvc}
D.~S. Cheng, M.~Cristani, M.~Stoppa, L.~Bazzani, and V.~Murino.
\newblock Custom pictorial structures for re-identification.
\newblock In {\em BMVC}, 2011.

\bibitem{Cui2007ChiEasyAlbum}
J.~Cui, F.~Wen, R.~Xiao, Y.~Tian, and X.~Tang.
\newblock Easyalbum: an interactive photo annotation system based on face
  clustering and re-ranking.
\newblock In {\em SIGCHI}, 2007.

\bibitem{Deng2009CvprImageNet}
J.~Deng, W.~Dong, R.~Socher, L.-J. Li, K.~Li, and L.~Fei-Fei.
\newblock {ImageNet: A Large-Scale Hierarchical Image Database}.
\newblock In {\em CVPR}, 2009.

\bibitem{Deng2014AcmPeta}
Y.~Deng, P.~Luo, C.~C. Loy, and X.~Tang.
\newblock Pedestrian attribute recognition at far distance.
\newblock In {\em ACMMM}, 2014.

\bibitem{Ding2015Arxiv}
C.~Ding and D.~Tao.
\newblock A comprehensive survey on pose-invariant face recognition.
\newblock {\em arXiv}, 2015.

\bibitem{Everingham2006Bmvc}
M.~Everingham, J.~Sivic, and A.~Zisserman.
\newblock Hello! my name is... buffy--automatic naming of characters in tv
  video.
\newblock In {\em BMVC}, 2006.

\bibitem{Everingham2009Ivc}
M.~Everingham, J.~Sivic, and A.~Zisserman.
\newblock Taking the bite out of automated naming of characters in tv video.
\newblock {\em IVC}, 2009.

\bibitem{Gallagher2008Cvpr}
A.~Gallagher and T.~Chen.
\newblock Clothing cosegmentation for recognizing people.
\newblock In {\em CVPR}, 2008.

\bibitem{Gallagher2007Cvpr}
A.~C. Gallagher and T.~Chen.
\newblock Using group prior to identify people in consumer images.
\newblock In {\em CVPR}, 2007.

\bibitem{Gandhi2013Cvpr}
V.~Gandhi and R.~Ronfard.
\newblock Detecting and naming actors in movies using generative appearance
  models.
\newblock In {\em CVPR}, 2013.

\bibitem{Garg2011Cvpr}
R.~Garg, S.~M. Seitz, D.~Ramanan, and N.~Snavely.
\newblock Where's waldo: matching people in images of crowds.
\newblock In {\em CVPR}, 2011.

\bibitem{Gong2014PersonReIdBook}
S.~Gong, M.~Cristani, S.~Yan, and C.~C. Loy.
\newblock {\em Person re-identification}.
\newblock Springer, 2014.

\bibitem{Guillaumin2009Iccv}
M.~Guillaumin, J.~Verbeek, and C.~Schmid.
\newblock Is that you? metric learning approaches for face identification.
\newblock In {\em ICCV}, 2009.

\bibitem{Hu2014Accvw}
Y.~Hu, D.~Yi, S.~Liao, Z.~Lei, and S.~Li.
\newblock Cross dataset person re-identification.
\newblock In {\em ACCV, workshop}, 2014.

\bibitem{Huang2007Lfw}
G.~B. Huang, M.~Ramesh, T.~Berg, and E.~Learned-Miller.
\newblock Labeled faces in the wild: A database for studying face recognition
  in unconstrained environments.
\newblock Technical report, UMass, 2007.

\bibitem{jia2014caffe}
Y.~Jia, E.~Shelhamer, J.~Donahue, S.~Karayev, J.~Long, R.~Girshick,
  S.~Guadarrama, and T.~Darrell.
\newblock Caffe: Convolutional architecture for fast feature embedding.
\newblock {\em arXiv}, 2014.

\bibitem{Krizhevsky2012Nips}
A.~Krizhevsky, I.~Sutskever, and G.~E. Hinton.
\newblock Imagenet classification with deep convolutional neural networks.
\newblock In {\em NIPS}, 2012.

\bibitem{Kumar2009Cvpr}
N.~Kumar, A.~C. Berg, P.~N. Belhumeur, and S.~K. Nayar.
\newblock Attribute and simile classifiers for face verification.
\newblock In {\em CVPR}, 2009.

\bibitem{Layne2012Bmvc}
R.~Layne, T.~M. Hospedales, S.~Gong, and Q.~Mary.
\newblock Person re-identification by attributes.
\newblock In {\em BMVC}, 2012.

\bibitem{Li2013Cvpr}
W.~Li and X.~Wang.
\newblock Locally aligned feature transforms across views.
\newblock In {\em CVPR}, 2013.

\bibitem{Li2014CvprDeepReID}
W.~Li, R.~Zhao, T.~Xiao, and X.~Wang.
\newblock Deepreid: Deep filter pairing neural network for person
  re-identification.
\newblock In {\em CVPR}, 2014.

\bibitem{Lin2010Eccv}
D.~Lin, A.~Kapoor, G.~Hua, and S.~Baker.
\newblock Joint people, event, and location recognition in personal photo
  collections using cross-domain context.
\newblock In {\em ECCV}. 2010.

\bibitem{Lu2014ArxivGaussianFace}
C.~Lu and X.~Tang.
\newblock Surpassing human-level face verification performance on lfw with
  gaussianface.
\newblock {\em arXiv}, 2014.

\bibitem{Mathialagan2015Arxiv}
C.~S. Mathialagan, A.~C. Gallagher, and D.~Batra.
\newblock Vip: Finding important people in images.
\newblock {\em arXiv}, 2015.

\bibitem{Mathias2014Eccv}
M.~Mathias, R.~Benenson, M.~Pedersoli, and L.~{Van Gool}.
\newblock Face detection without bells and whistles.
\newblock In {\em ECCV}, 2014.

\bibitem{Oliva2001IjcvGist}
A.~Oliva and A.~Torralba.
\newblock Modeling the shape of the scene: A holistic representation of the
  spatial envelope.
\newblock {\em IJCV}, 2001.

\bibitem{Schroff2015ArxivFaceNet}
F.~Schroff, D.~Kalenichenko, and J.~Philbin.
\newblock Facenet: A unified embedding for face recognition and clustering.
\newblock {\em arXiv}, 2015.

\bibitem{Shi2013Iccv}
J.~Shi, R.~Liao, and J.~Jia.
\newblock Codel: A human co-detection and labeling framework.
\newblock In {\em ICCV}, 2013.

\bibitem{Stone2008Cvprw}
Z.~Stone, T.~Zickler, and T.~Darrell.
\newblock Autotagging facebook: Social network context improves photo
  annotation.
\newblock In {\em CVPR workshops}, 2008.

\bibitem{Sun2014ArxivDeepId2plus}
Y.~Sun, X.~Wang, and X.~Tang.
\newblock Deeply learned face representations are sparse, selective, and
  robust.
\newblock {\em arXiv}, 2014.

\bibitem{Taigman2014CvprDeepFace}
Y.~Taigman, M.~Yang, M.~Ranzato, and L.~Wolf.
\newblock Deepface: Closing the gap to human-level performance in face
  verification.
\newblock In {\em CVPR}, 2014.

\bibitem{Yi2014Arxiv}
D.~Yi, Z.~Lei, and S.~Z. Li.
\newblock Deep metric learning for practical person re-identification.
\newblock {\em arXiv}, 2014.

\bibitem{Yi2014ArxivLearningFace}
D.~Yi, Z.~Lei, S.~Liao, and S.~Z. Li.
\newblock Learning face representation from scratch.
\newblock {\em arXiv}, 2014.

\bibitem{Zhang2015CvprPiper}
N.~Zhang, M.~Paluri, Y.~Taigman, R.~Fergus, and L.~Bourdev.
\newblock Beyond frontal faces: Improving person recognition using multiple
  cues.
\newblock In {\em CVPR}, 2015.

\bibitem{Zhao2013IccvSalienceMatching}
R.~Zhao, W.~Ouyang, and X.~Wang.
\newblock Person re-identification by salience matching.
\newblock In {\em ICCV}, 2013.

\bibitem{Zhou2014NipsPlaces}
B.~Zhou, A.~Lapedriza, J.~Xiao, A.~Torralba, and A.~Oliva.
\newblock {Learning Deep Features for Scene Recognition using Places Database.}
\newblock {\em NIPS}, 2014.

\bibitem{Zhou2015ArxivNaiveDeepFace}
E.~Zhou, Z.~Cao, and Q.~Yin.
\newblock Naive-deep face recognition: Touching the limit of lfw benchmark or
  not?
\newblock {\em arXiv}, 2015.

\bibitem{Zhu2013Iccv}
Z.~Zhu, P.~Luo, X.~Wang, and X.~Tang.
\newblock Deep learning identity-preserving face space.
\newblock In {\em ICCV}, 2013.

\end{thebibliography}

\end{otherlanguage}

\clearpage{}

\appendix

\part*{{\LARGE{}Appendix}}

\section{\label{sec:Content}Content}

This appendix provides additional qualitative and quantitative details
of the experiments and results discussed in the main paper. It includes
visualisations of newly proposed splits (\S\ref{sec:split-examples}),
success and failure examples of our systems (\S\ref{sec:success-cases},\ref{sec:failure-modes}),
detailed validation and test set tables (\S\ref{sec:detailed-results}),
and other technical details for the experiments (\S\ref{sec:detector-details},
\ref{sec:fine-tuning}, \ref{sec:svm-vs-nn}, \ref{sec:Attributes-supp}).

\section{\label{sec:split-examples}More examples of splits}

\begin{figure*}
\begin{centering}
\hspace*{\fill}\includegraphics[width=2\columnwidth]{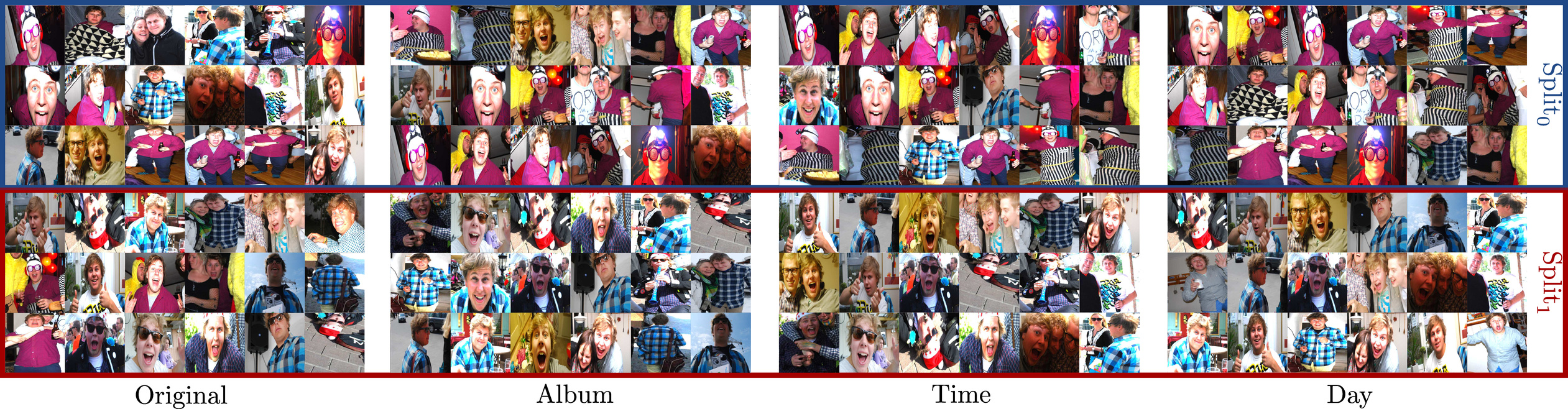}\hspace*{\fill}
\par\end{centering}

\begin{centering}
\vspace{0.5em}

\par\end{centering}

\begin{centering}
\hspace*{\fill}\includegraphics[width=2\columnwidth]{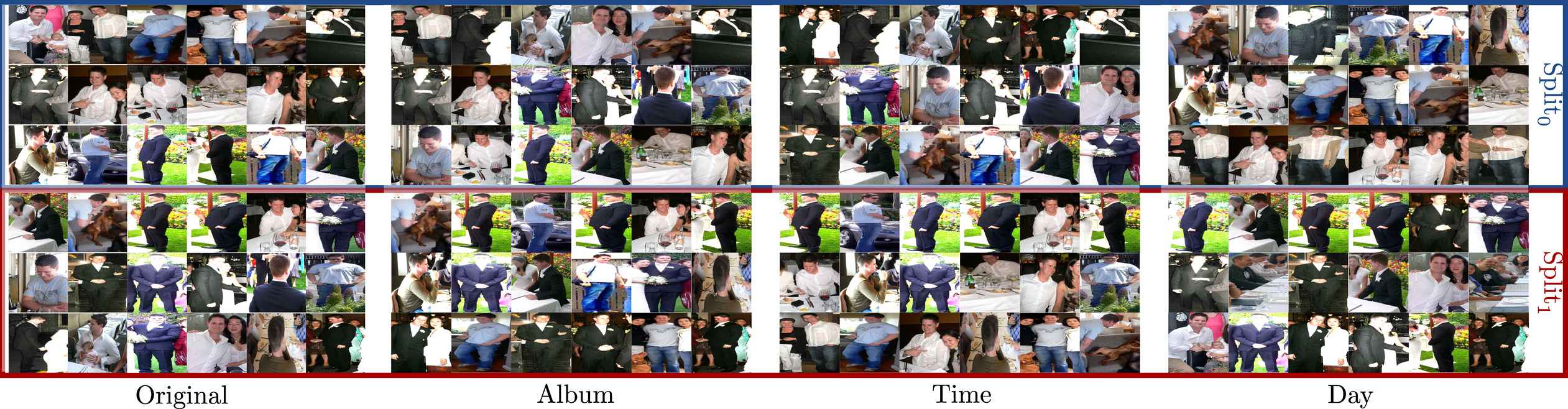}\hspace*{\fill}
\par\end{centering}

\begin{centering}
\vspace{0.5em}

\par\end{centering}

\begin{centering}
\hspace*{\fill}\includegraphics[width=2\columnwidth]{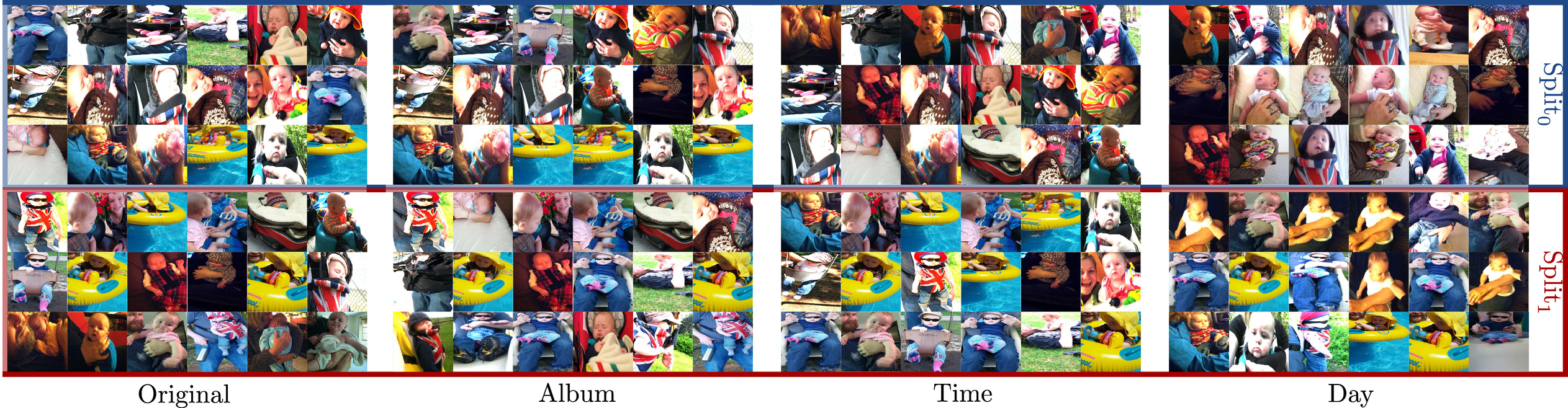}\hspace*{\fill}
\par\end{centering}

\begin{centering}
\vspace{0em}

\par\end{centering}

\protect\caption{\label{fig:splits-visualisation-appendix}Example of different split
types over three identities.}
\end{figure*}

We provide more examples of the proposed split on the PIPA dataset
(Figure \ref{fig:splits-visualisation-appendix}). The separation
of appearances across $\mbox{split}_{0/1}$ becomes clearer as we
shift from the Original to Day splits.

\section{\label{sec:success-cases}More success and failure examples}

\begin{figure*}
\begin{centering}
\hspace*{\fill}\includegraphics[width=0.85\columnwidth]{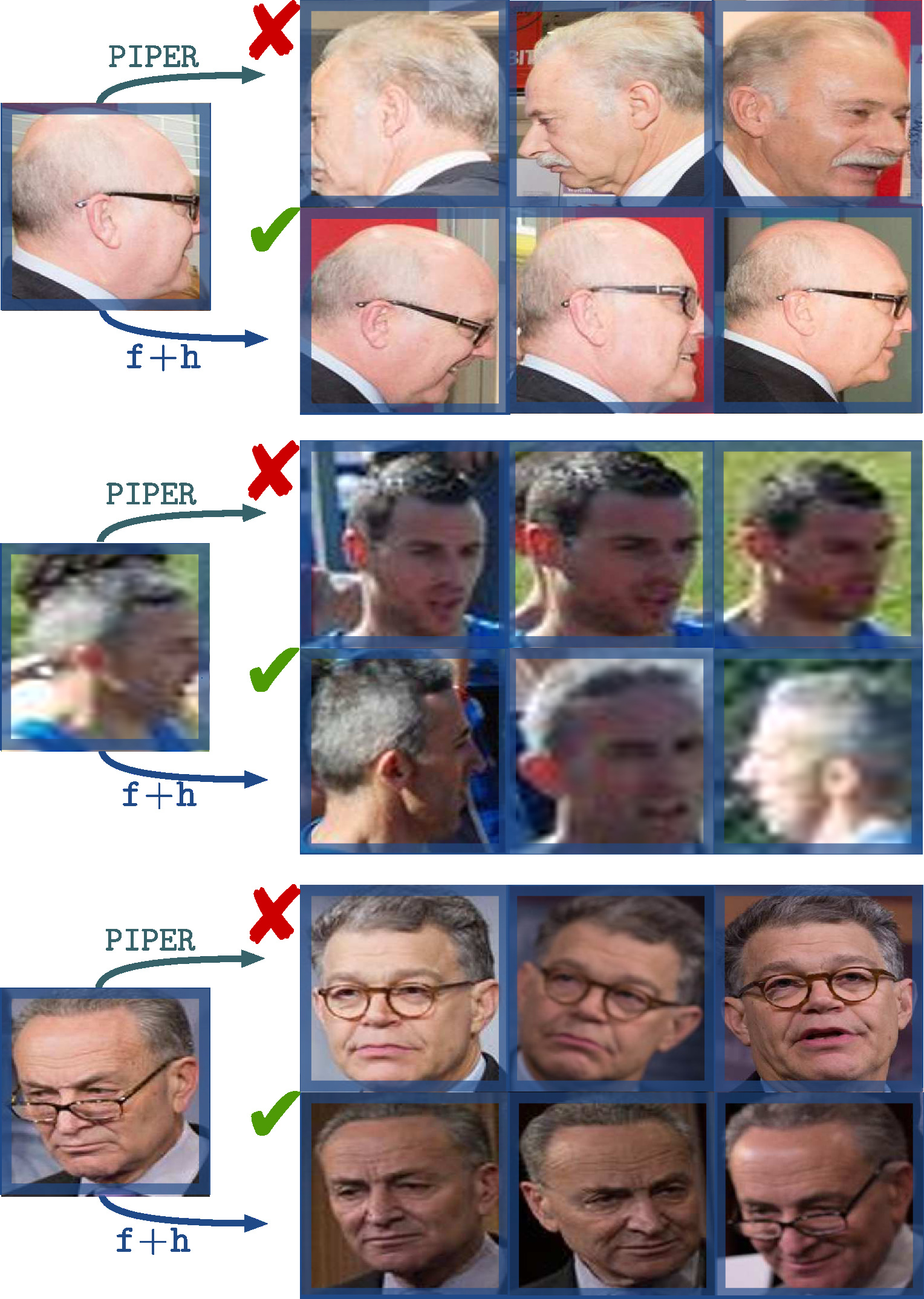}\hspace*{\fill}\includegraphics[width=0.85\columnwidth]{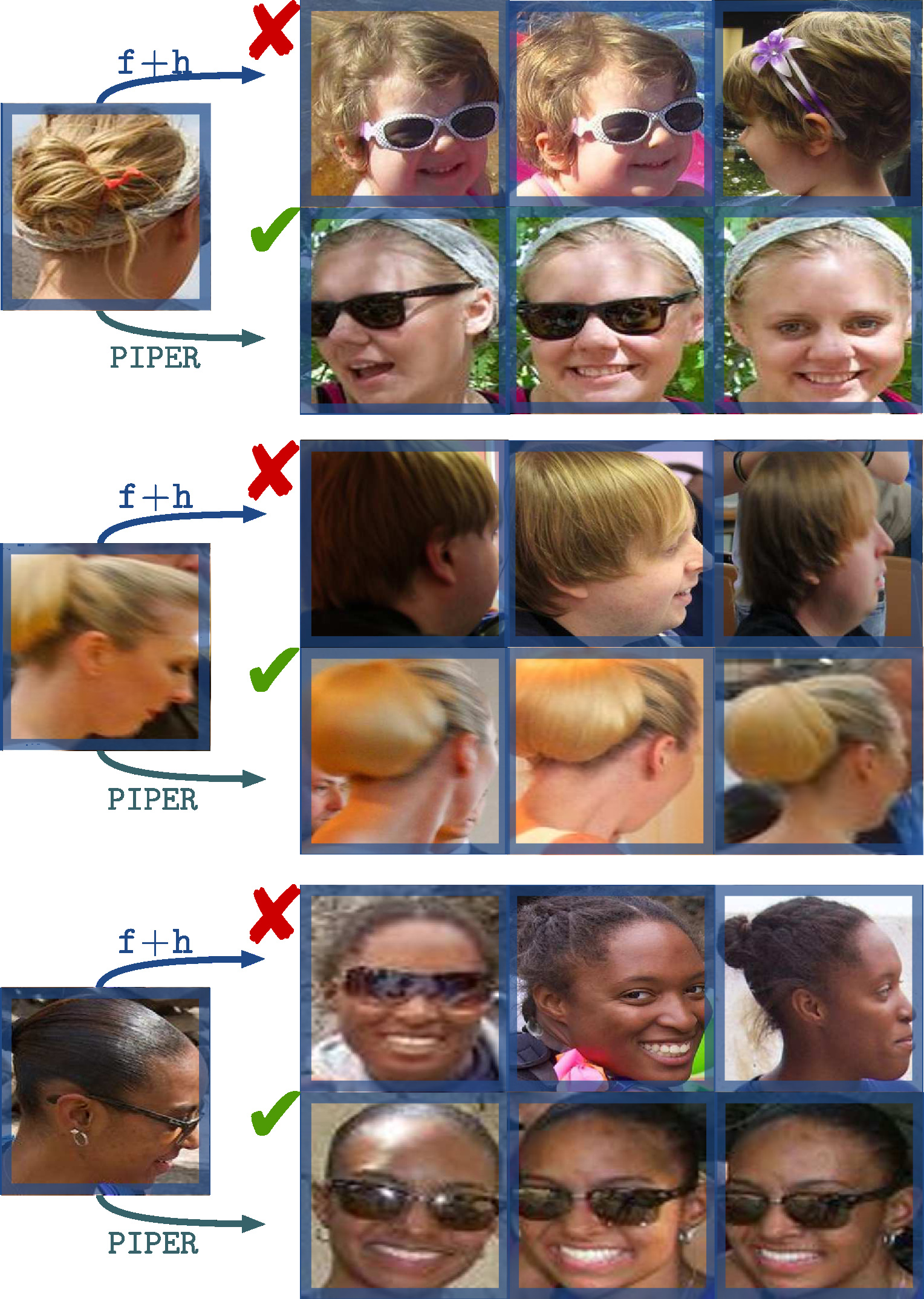}\hspace*{\fill}
\par\end{centering}

\begin{centering}
\vspace{0em}

\par\end{centering}

\protect\caption{\label{fig:success-failure-head}Success and failure cases of $\texttt{f}\negthinspace+\negthinspace\texttt{h}$
under the Original split. Left column, $\texttt{PIPER}$ fails, $\texttt{f}\negthinspace+\negthinspace\texttt{h}$
recognizes correctly. Right column, shows the inverse case.}
\end{figure*}

\begin{figure*}
\begin{centering}
\hspace*{\fill}\includegraphics[width=0.85\columnwidth]{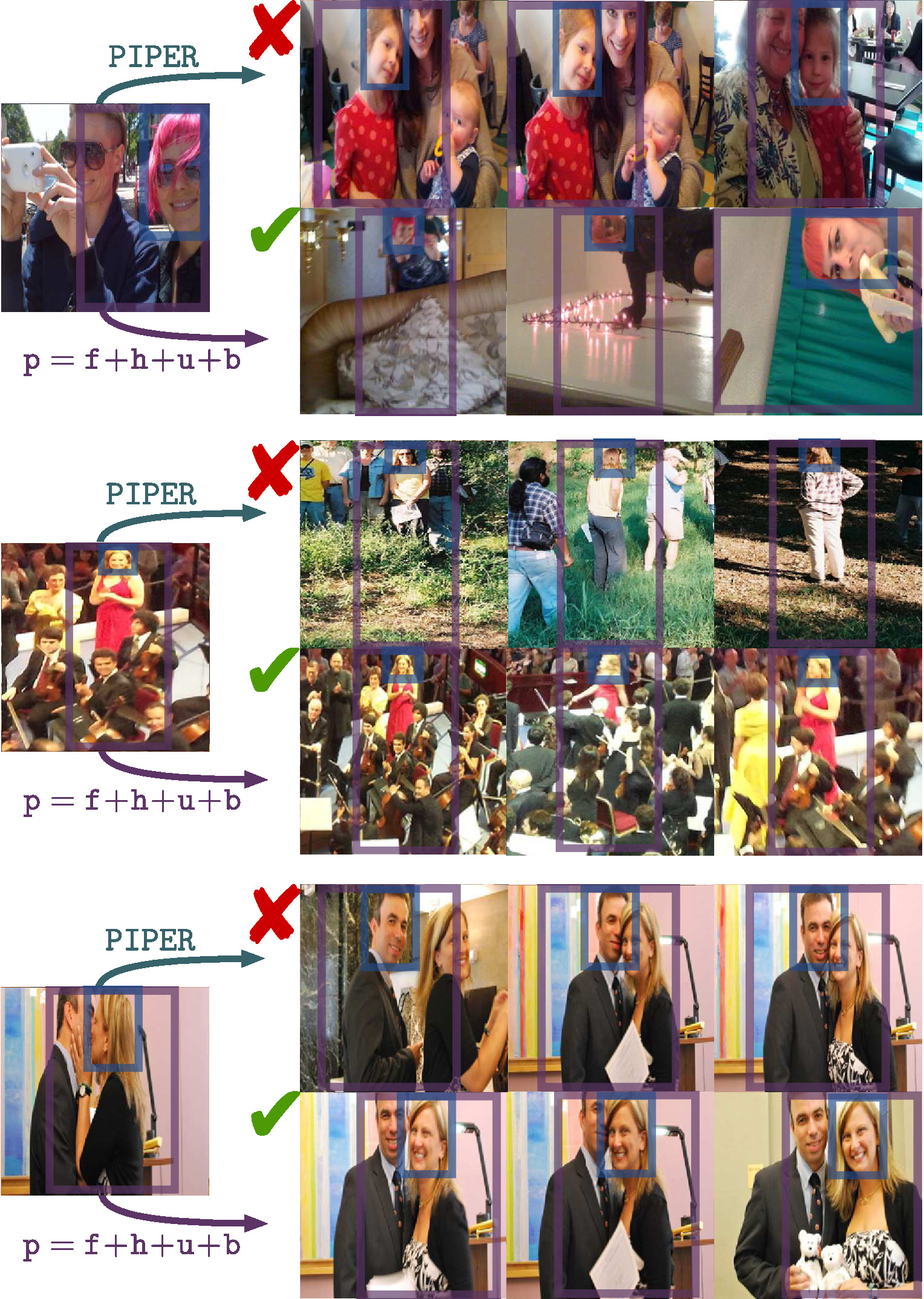}\hspace*{\fill}\includegraphics[width=0.85\columnwidth]{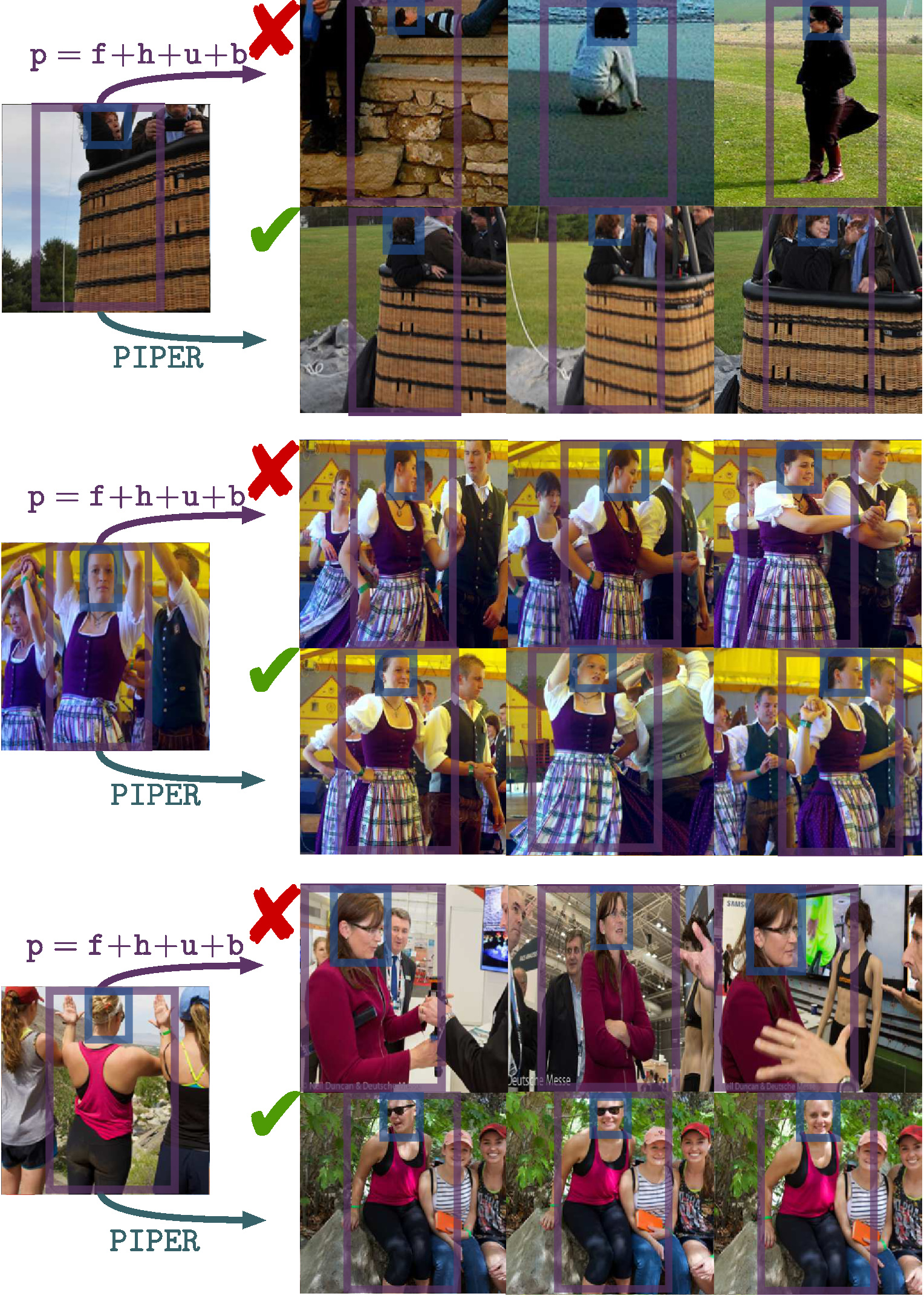}\hspace*{\fill}
\par\end{centering}

\begin{centering}
\vspace{0em}

\par\end{centering}

\protect\caption{\label{fig:success-failure-person}Success and failure cases of $\texttt{p}=\texttt{f}\negthinspace+\negthinspace\texttt{h}\negthinspace+\negthinspace\texttt{u}\negthinspace+\negthinspace\texttt{b}$
under the Original split. }
\end{figure*}

\begin{figure*}
\begin{centering}
\hspace*{\fill}\includegraphics[width=1\columnwidth]{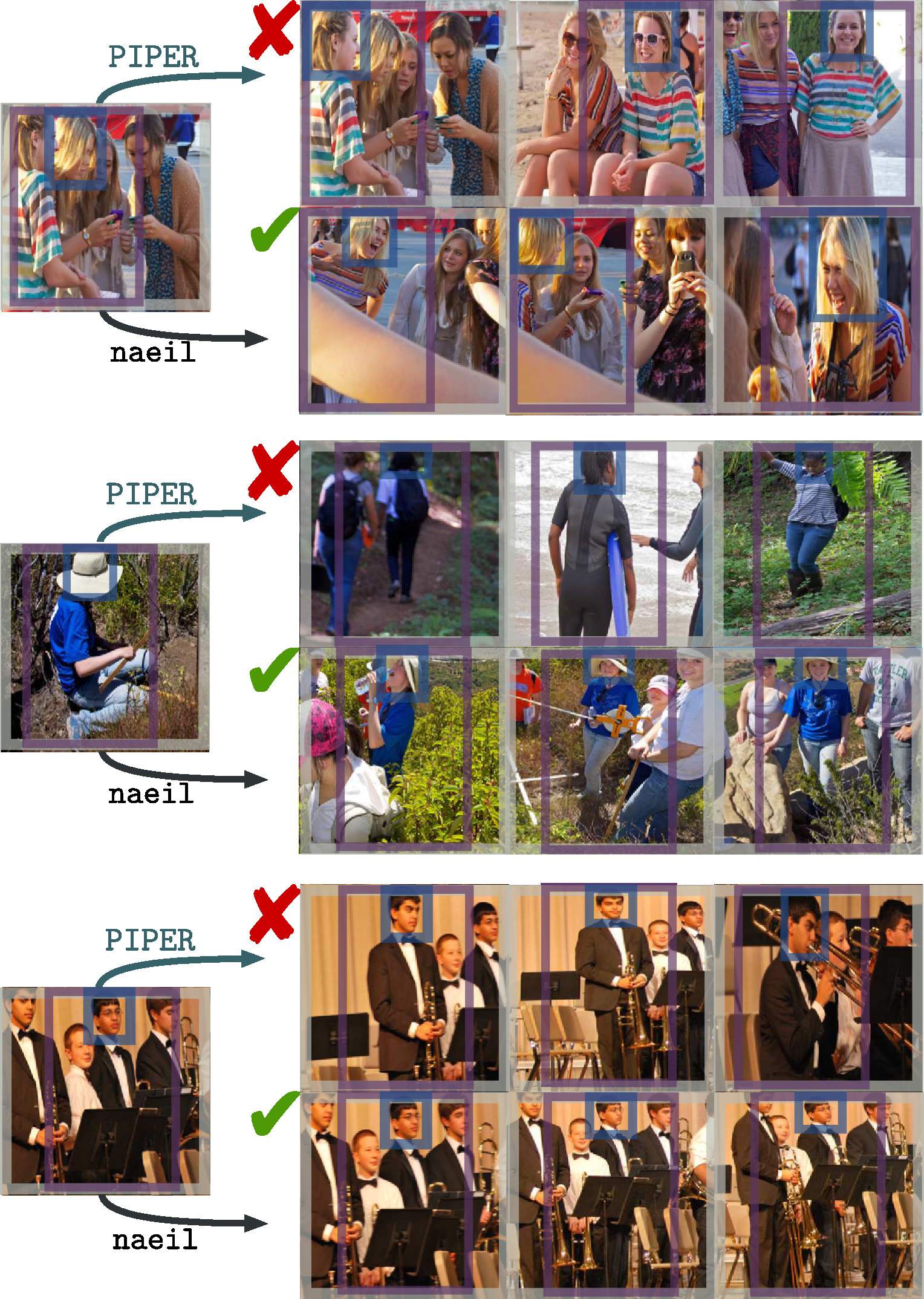}\hspace*{\fill}\includegraphics[width=1\columnwidth]{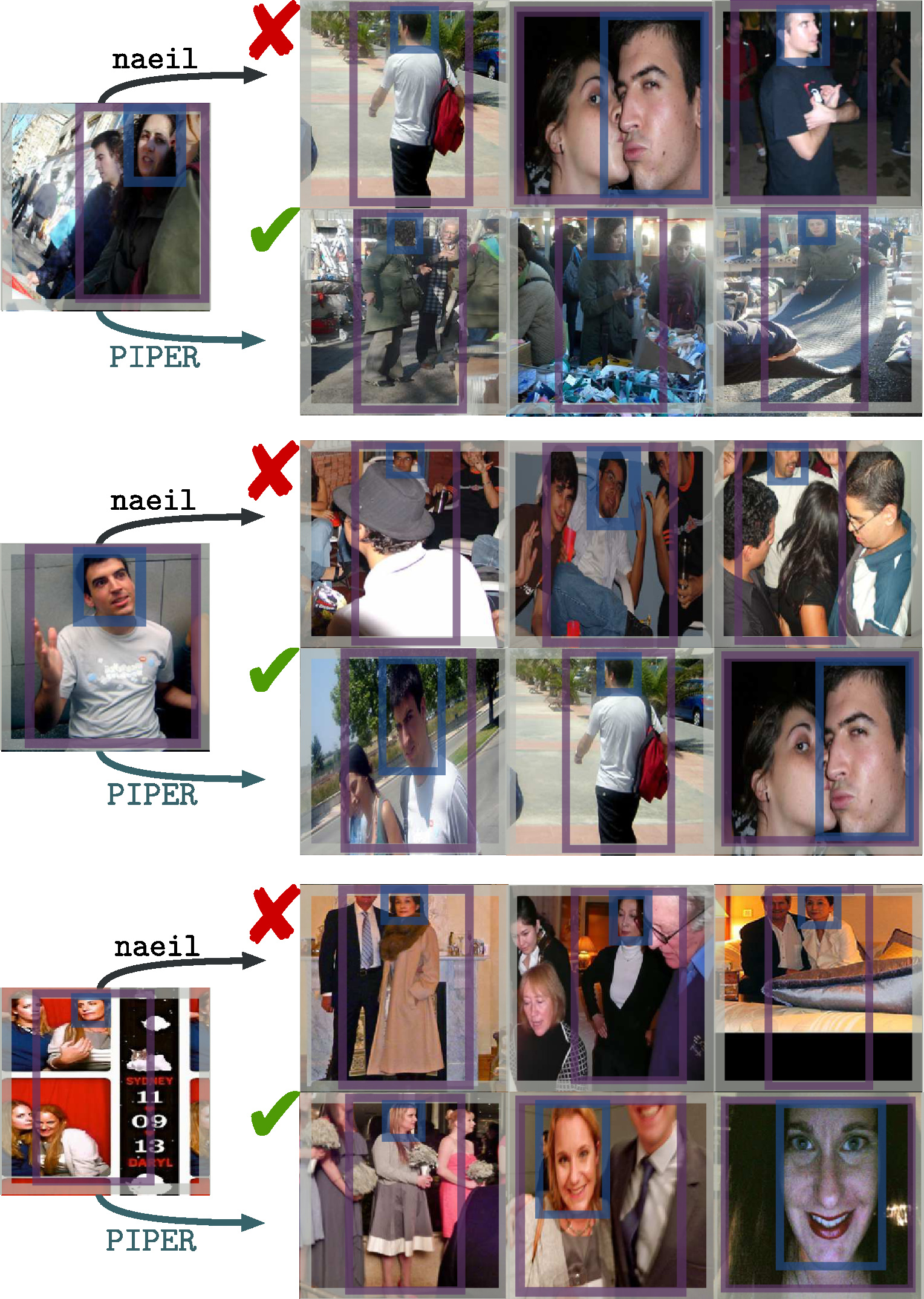}\hspace*{\fill}
\par\end{centering}

\begin{centering}
\vspace{0em}

\par\end{centering}

\protect\caption{\label{fig:success-failure-naeil}Success and failure cases of $\texttt{naeil}$
under the Original split. }
\end{figure*}

We provide additional qualitative examples of success and failure
cases of our systems. Figures \ref{fig:success-failure-head} to \ref{fig:success-failure-naeil}
show test instances (single images on the left) and training instances
(triple images on the right) of the identity that the system predicted.
The triple training instances are ordered in the nearest $L_{2}$
feature distance from the test image. The ticks and crosses denote
whether the system's prediction was correct or not. Note the symmetry
of the left and right columns: left columns are cases that our systems
($\texttt{f}\negthinspace+\negthinspace\texttt{h}$, $\texttt{p}=\texttt{f}\negthinspace+\negthinspace\texttt{h}\negthinspace+\negthinspace\texttt{u}\negthinspace+\negthinspace\texttt{b}$,
$\texttt{naeil}$) correctly predicted while the $\texttt{PIPER}$
did not, and the right columns present the reversed case.

\begin{figure*}
\begin{centering}
\hspace*{\fill}\includegraphics[bb=0bp 0bp 367bp 757bp,width=1\columnwidth]{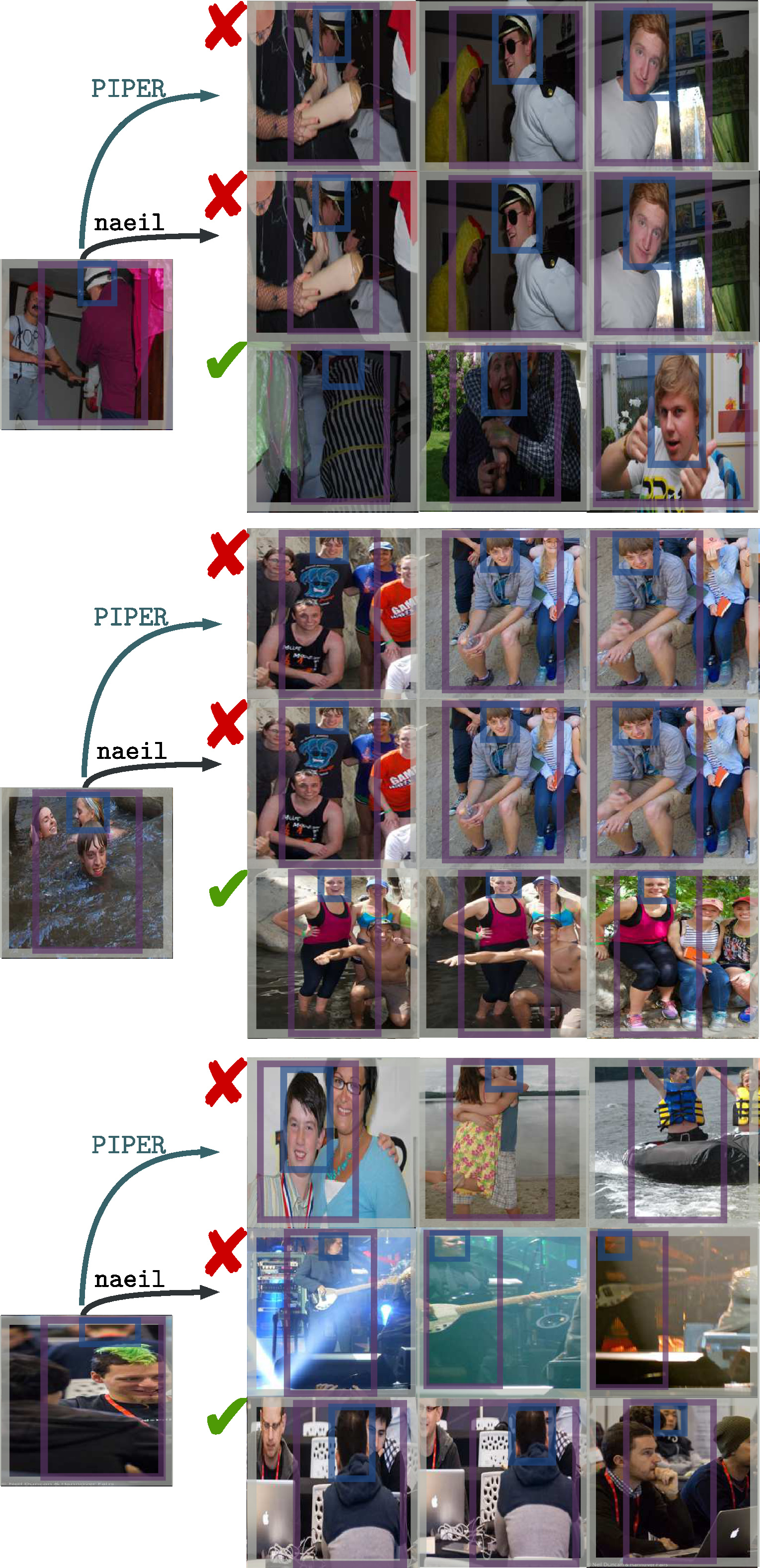}\hspace*{\fill}\includegraphics[bb=0bp 0bp 367bp 757bp,width=1\columnwidth]{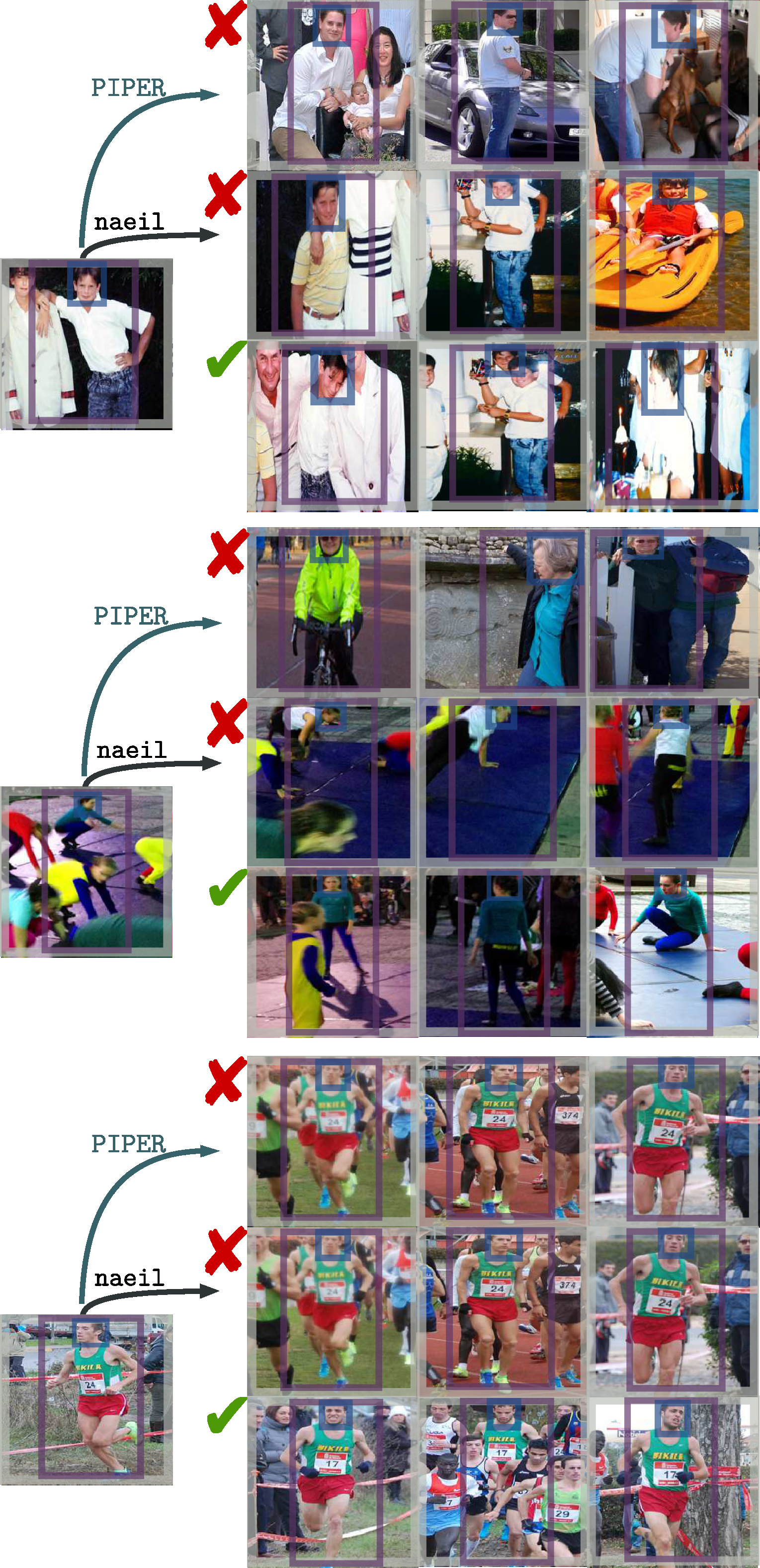}\hspace*{\fill}
\par\end{centering}

\begin{centering}
\vspace{0em}

\par\end{centering}

\protect\caption{\label{fig:fail-fail}Failure examples of both $\texttt{naeil}$ and
$\texttt{PIPER}$ under the Original split. }
\end{figure*}

We also provide examples where neither $\texttt{naeil}$ nor $\texttt{PIPER}$
correctly predicted (Figure \ref{fig:fail-fail}), which correspond
to 9.35\% of the whole test set. One can observe inter-personal confusion
due to similar clothing and background similarity (left top/middle,
right top/middle), severe occlusions of body regions in the test image
(left top/middle/bottom), and  an annotation error (right bottom;
note the marathoner's front number).

\section{\label{sec:failure-modes}Failure modes}

\begin{figure}
\begin{centering}
\hspace*{\fill}\includegraphics[width=0.8\columnwidth]{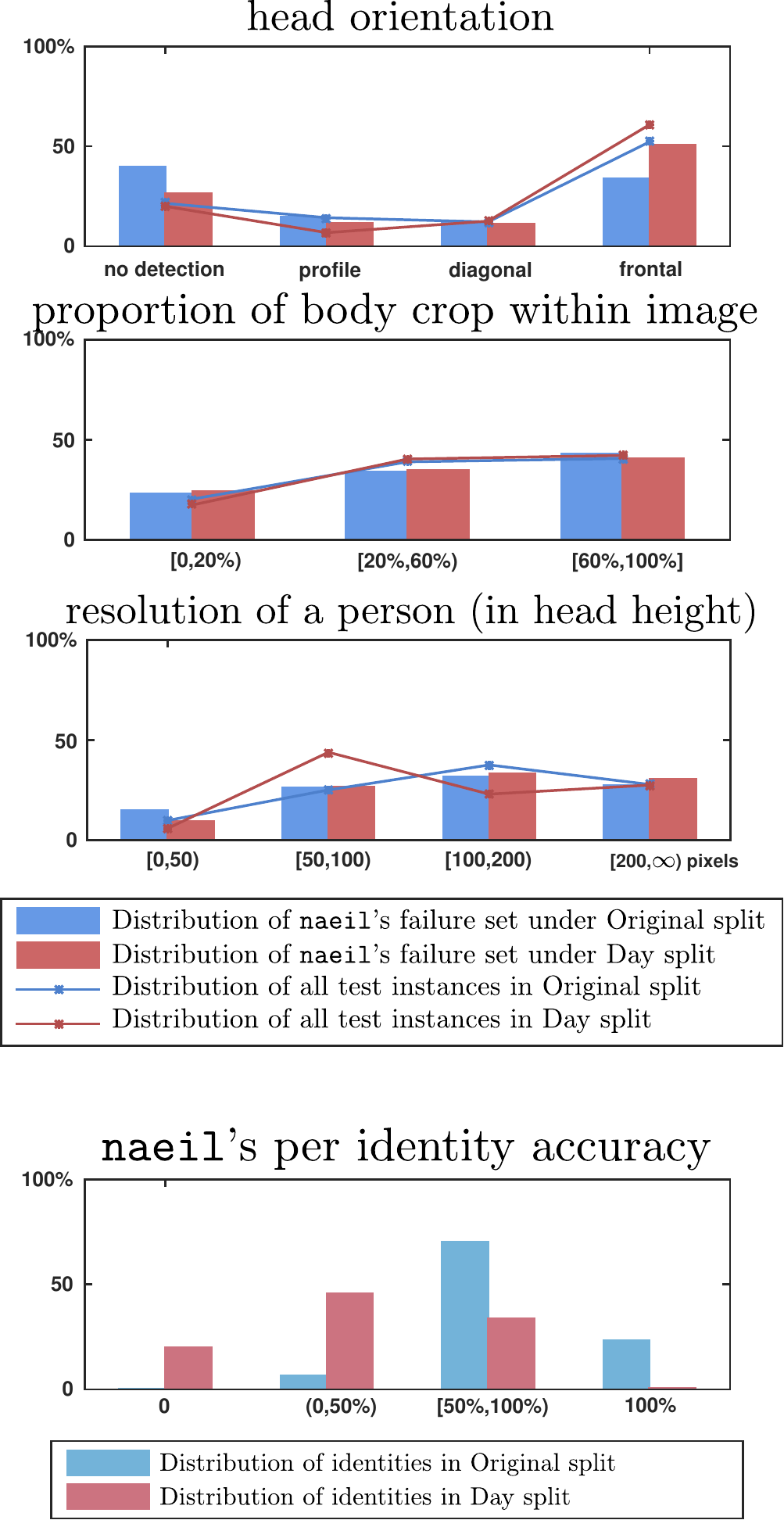}\hspace*{\fill}
\par\end{centering}

\begin{centering}
\vspace{0em}

\par\end{centering}

\protect\caption{\label{fig:failure-factors}Top three: distribution of instances with
respect to failure factors. Bottom: distribution of identities according
to $\texttt{naeil}$'s performance for each identity. }
\end{figure}

We also provide an auxiliary visualisation to the observations in
\S\ref{sub:Analysis-of-failure}. The top three plots in Figure \ref{fig:failure-factors}
show the distribution of $\texttt{naeil}$'s failure cases with respect
to three different factors: head orientation, resolution, and the
body crop truncation. The bottom plot analyses per identity accuracy
of $\texttt{naeil}$.

\paragraph{Head orientation}

For head orientation, we see that indeed failure cases have a greater
proportion of non-frontal faces compared to the entire test set distribution.
However, it can also be deduced that the failure cases are less correlated
to the head orientations in the Day split setting, from the fact that
the failure distribution in the Day split deviates less from the entire
population's distribution than does the Original split counterpart.

\paragraph{Body crop truncation}

In the body crop truncation plot, we observe a homogeneity between
Original and Day split distributions, and that having less image content
in the body crop is indeed detrimental to the recognition in both
Original and Day splits.

\paragraph{Resolution}

The resolution plot shows how the resolution of a person instance
is related to $\texttt{naeil}$'s ability to recognise the person.
Note that head height was measured, since all different body crops
(which excludes scene $\texttt{s}$) are proportional to the head
size. Under the Day split, $\texttt{naeil}$ has greater proportion
of medium resolution heads ($[100,200)$ pixels) than lower resolution
heads, while the entire population has greater proportion of lower
resolution heads. In other words, resolution is not positively correlated
with naeil's performance under the Day split. Hence, for example,
picturing a person at a closer distance is not likely to greatly improve
recognition across days.

\paragraph{Per identity accuracy}

The final plot shows the $\texttt{naeil}$'s per identity performance.
Note the increase in the proportion of never-identified individuals
(leftmost bin) from the Original split to the Day split. This suggests
that under the Day split there exist a meaningful number of identities
which $\texttt{naeil}$ cannot currently handle. 
\begin{figure}
\begin{centering}
\includegraphics[width=0.9\columnwidth]{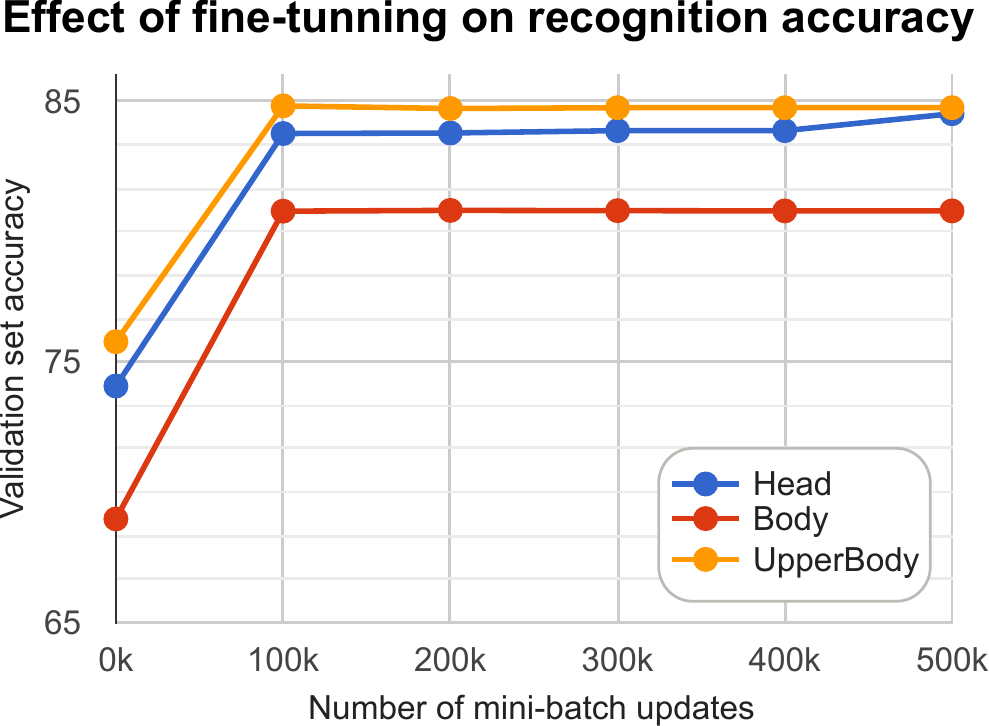}\vspace{-0.5em}

\par\end{centering}

\protect\caption{\label{fig:fine-tuning-effect}Validation set performance of different
cues, as a function of the fine-tuning duration.}
\end{figure}

\section{\label{sec:detector-details}Face detector details}

\begin{figure*}
\begin{centering}
\hspace*{\fill}\subfloat[$-90\degree$]{\protect\begin{centering}
\protect\includegraphics[width=0.15\textwidth]{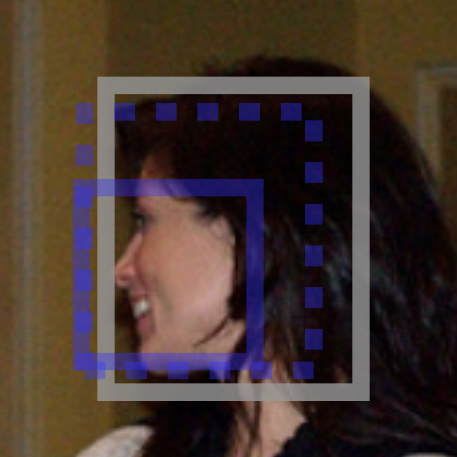}\hspace*{0.3em}\protect\includegraphics[width=0.15\textwidth]{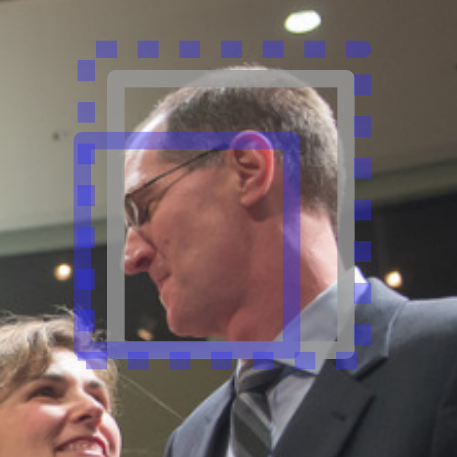}\hspace*{0.3em}\protect\includegraphics[width=0.15\textwidth]{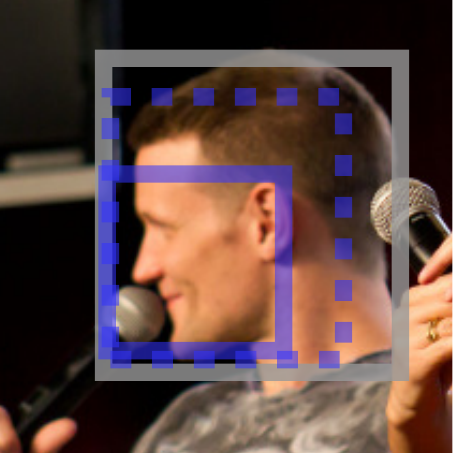}\protect
\par\end{centering}

}\hspace*{\fill}\subfloat[$+90\degree$]{\protect\centering{}\protect\includegraphics[width=0.15\textwidth]{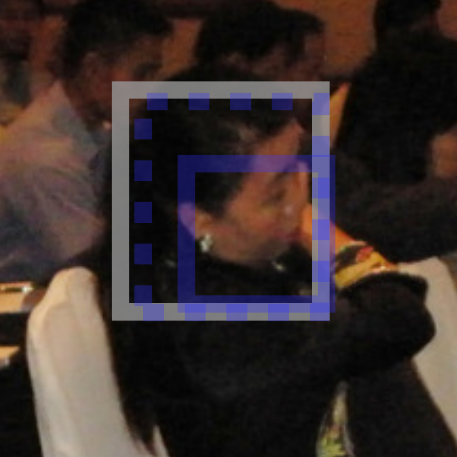}\hspace*{0.3em}\protect\includegraphics[width=0.15\textwidth]{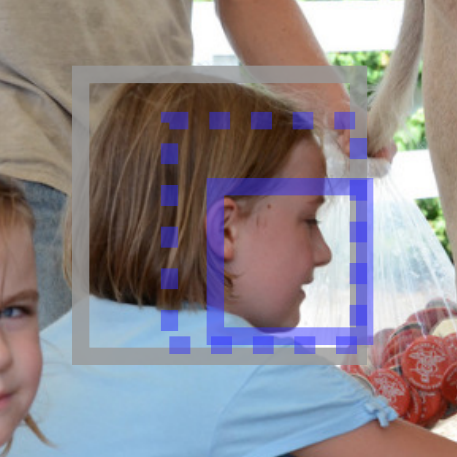}\hspace*{0.3em}\protect\includegraphics[width=0.15\textwidth]{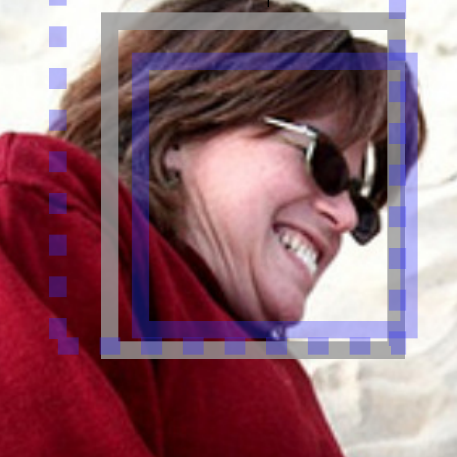}\protect}\hspace*{\fill}
\par\end{centering}

\begin{centering}
\hspace*{\fill}\subfloat[$-45\degree$]{\protect\begin{centering}
\protect\includegraphics[width=0.15\textwidth]{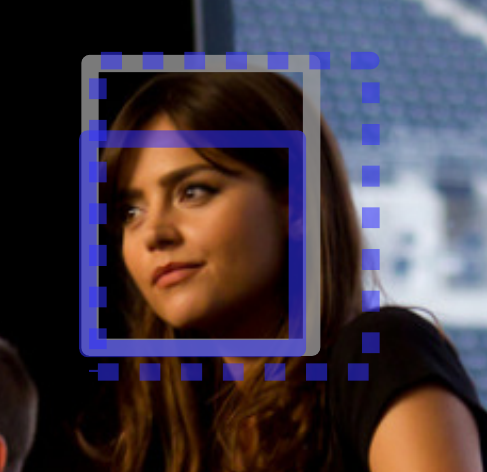}\hspace*{0.3em}\protect\includegraphics[width=0.15\textwidth]{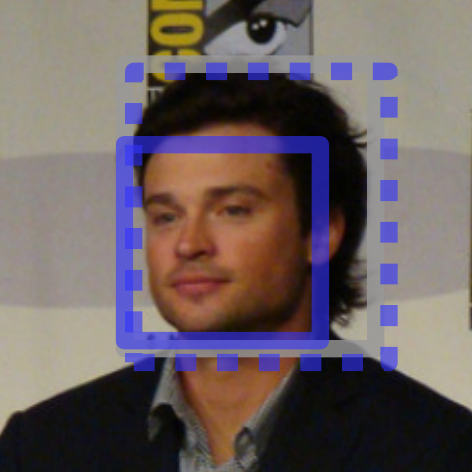}\hspace*{0.3em}\protect\includegraphics[width=0.15\textwidth]{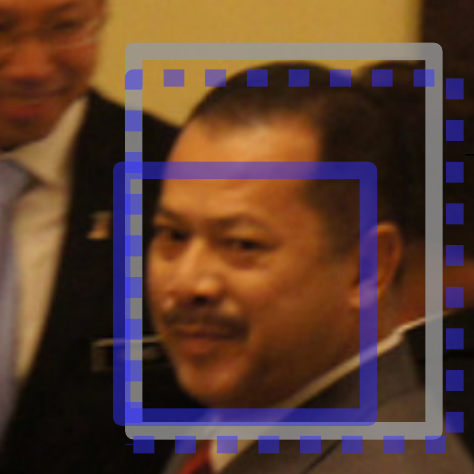}\protect
\par\end{centering}

}\hspace*{\fill}\subfloat[$+45\degree$]{\protect\centering{}\protect\includegraphics[width=0.15\textwidth]{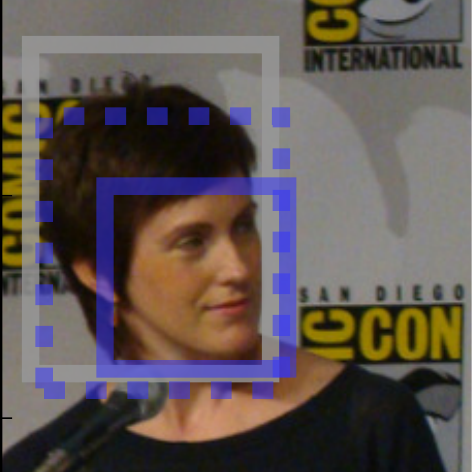}\hspace*{0.3em}\protect\includegraphics[width=0.15\textwidth]{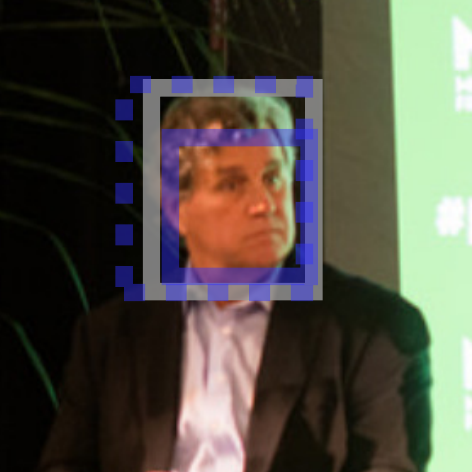}\hspace*{0.3em}\protect\includegraphics[width=0.15\textwidth]{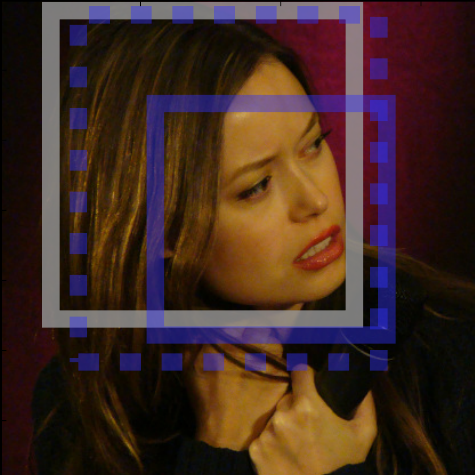}\protect}\hspace*{\fill}
\par\end{centering}

\hspace*{\fill}\subfloat[$\pm0\degree$]{\protect\centering{}\protect\includegraphics[width=0.15\textwidth]{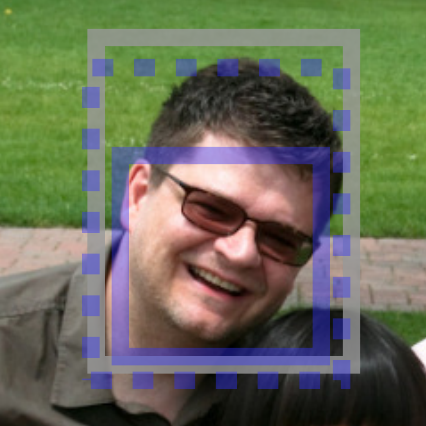}\hspace*{0.3em}\protect\includegraphics[width=0.15\textwidth]{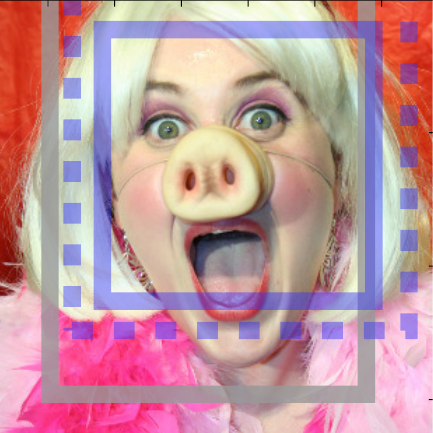}\hspace*{0.3em}\protect\includegraphics[width=0.15\textwidth]{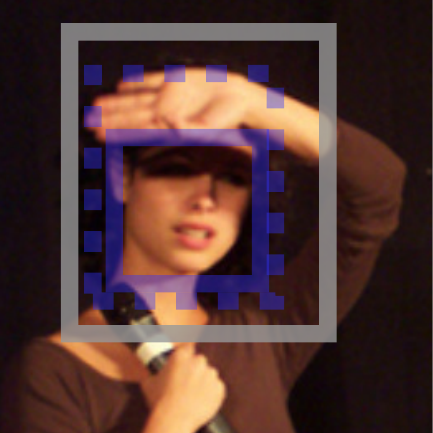}\protect}\hspace*{\fill}\subfloat[Missing detections]{\protect\centering{}\protect\includegraphics[width=0.15\textwidth]{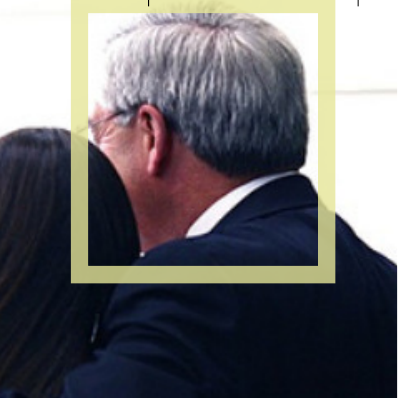}\hspace*{0.3em}\protect\includegraphics[width=0.15\textwidth]{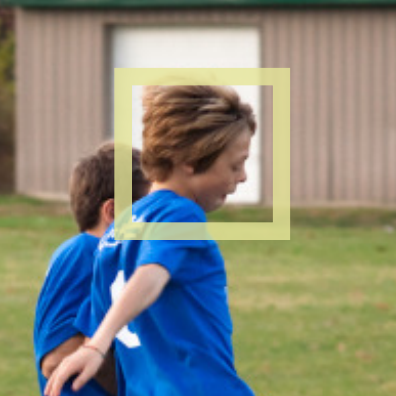}\hspace*{0.3em}\protect\includegraphics[width=0.15\textwidth]{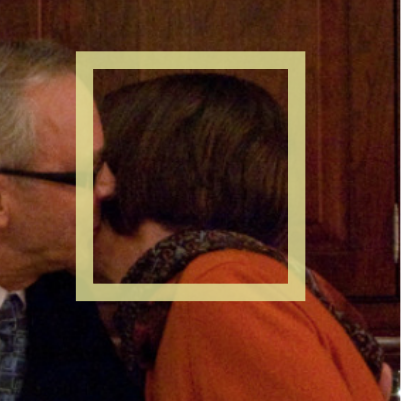}\protect}\hspace*{\fill}

\hspace*{\fill}\subfloat[Legend]{\protect\centering{}\hspace{1.6em}\protect\includegraphics[height=0.15\textwidth]{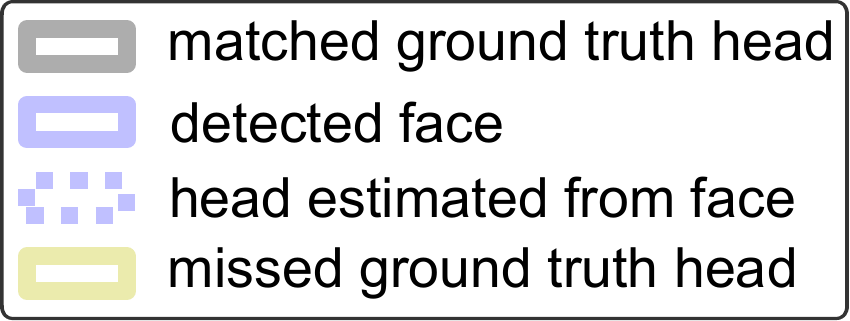}\hspace{1.6em}\protect}\hspace*{\fill}\subfloat[Detected heads, but wrong orientation estimate]{\protect\centering{}\protect\includegraphics[width=0.15\textwidth]{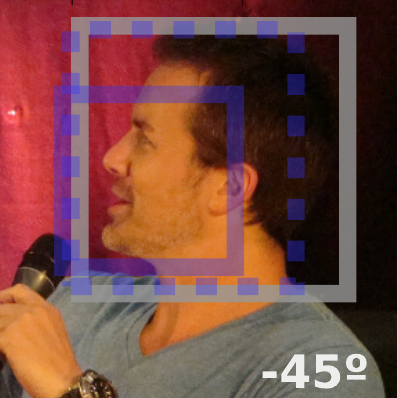}\hspace*{0.3em}\protect\includegraphics[width=0.15\textwidth]{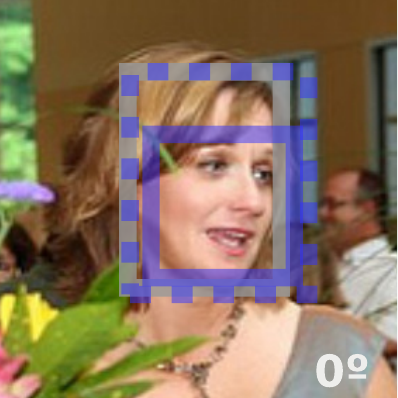}\hspace*{0.3em}\protect\includegraphics[width=0.15\textwidth]{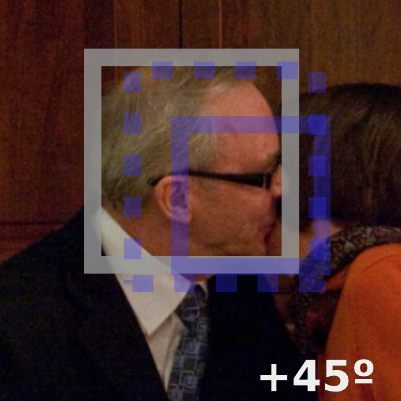}\protect}\hspace*{\fill}\protect\caption{\label{fig:face-detection-examples}Examples results from the face
detector (PIPA validation set), and estimated head boxes. }
\end{figure*}

For face detection we use the state of the art DPM detector from \cite{Mathias2014Eccv}.
This detector is trained on $\sim\negmedspace15\mbox{k}$ faces from
the AFLW database, and is composed of $6$ components which give a
rough indication of face orientation: $\pm0\degree$ (frontal), $\pm45\degree$
(diagonal left and right), and $\pm90\degree$ (side views). Figure
\ref{fig:face-detection-examples} shows example face detections on
the PIPA dataset. It shows detections, the estimated orientation,
the regressed head bounding box, the corresponding ground truth head
box, and some failure modes. Faces corresponding to $\pm0\degree$
are considered frontal, and all others ($\pm45\degree$, $\pm90\degree$,
and non-detected) are considered non-frontal. No ground truth is available
to evaluate the face orientation estimation, except a few mistakes,
the $\pm0\degree$ components seems a rather reliable estimators (while
more confusion is observed between $\pm45\degree$/$\pm90\degree$).

\section{\label{sec:detailed-results}Detailed results}

\subsection{\label{sec:detailed-validation-results}Detailed validation set results}

See Table \ref{tab:validation-set-regions-accuracy-appendix} for
detailed results on the validation set. It also shows the increase
in performance as we zoom out/in from the face($\texttt{f}$)/scene($\texttt{s}$).
When we zoom out, we already gain most of the identity information
from the face to upper body regions, and the rest contributes only
marginally. As we zoom in, a rather gradual improvement is observed.
It is notable, however, that $\texttt{s}\negthinspace+\negthinspace\texttt{b}$
(82.16\%) is already almost as effective as the two part counterpart
in the zoom out scenario, $\texttt{f}\negthinspace+\negthinspace\texttt{h}$
(84.80\%).

\subsection{\label{sec:detailed-test-results}Detailed test set results}

See Table \ref{tab:splits-accuracy} for the test set results for
different experimental setups. Note that the addition of external
data, including the attribute cues (PIPA attributes and PETA attributes)
and large face databases (CACD and CASIA), is especially effective
in the Day split setting: from 42.31\% by $\texttt{\ensuremath{\mbox{\ensuremath{\texttt{P}}}_{s}}}=\texttt{P}\negthinspace+\negthinspace\texttt{s}$
to 46.54\% by \texttt{$\texttt{naeil}=\texttt{\ensuremath{\mbox{\ensuremath{\texttt{P}}}_{s}}+\texttt{E}}$.}

\section{\label{sec:fine-tuning}How much fine-tuning ?}

\paragraph{Task}

Unless otherwise stated, we fine-tune the ImageNet pre-trained AlexNet
\cite{Krizhevsky2012Nips} on the PIPA person recognition train set.
The initial weights of the AlexNet are obtained by training on ImageNet
for objects classification, and are further optimised by training
with different region crops of PIPA train set images for the identity
classification task.

\paragraph{Number of iterations}

Figure \ref{fig:fine-tuning-effect} verifies that $300\mbox{k}$
iterations with mini batch size $50$ gives maximal, or close to maximal,
performance for most cues. In fact the plateau is reached already
at $100\mbox{k}$, but we use $300\mbox{k}$ as precaution. Note that
we do not observe any over-fitting behaviour. \\
In the main paper we report the results of fine-tuning for the scene
region $\texttt{s}$, this is the only region that does not show a
large gain due to fine-tuning.\\
The results of Figure \ref{fig:fine-tuning-effect} are obtained by
training and testing SVMs on the PIPA validation set original splits,
using AlexNet features obtained via fine-tuning on the PIPA training
set.

\paragraph{Implementation details}

Our implementation uses the Caffe library \cite{jia2014caffe}\footnote{https://github.com/BVLC/caffe},
and the provided AlexNet model \texttt{bvlc\_alexnet.caffemodel}.\\
For fine-tuning, we use the following training configurations parameters:
\begin{verbatim}
(prototxt for solver configuration)
base_lr: 0.0001 
lr_policy: "step" 
gamma: 0.1 
stepsize: 50000
momentum: 0.9 
weight_decay: 0.0005

(prototxt for net specification)
batch_size: 50
\end{verbatim}
\selectlanguage{british}%
Regarding the per-identity SVMs, we fix the SVM parameter $C$ at
$1$ throughout the paper. Preliminary experiments indicated that
this was not a sensitive parameter.

\section{\label{sec:svm-vs-nn}SVM versus NN}

Table \ref{tab:svm-vs-nn} compares the validation set accuracy of
different cue combinations, when using (per-identity) SVM classifiers
or a nearest neighbour (NN) classifier.

The results show that using an SVM per identity is consistently better
than a naive nearest neighbour classifier.

\section{\label{sec:Attributes-supp}Attributes}

Table \ref{tab:attributes-details} shows the definitions of attribute
classes that we annotated on PIPA head crops. We did not annotate
attributes for identities (1) whose appearances are indecisive for
attribute classification (e.g. gender), and (2) whose attributes change
in PIPA (e.g. sunglasses). We will release the annotations.

For upper body crops, we use the PETA dataset \cite{Deng2014AcmPeta}
and five selected binary attribute annotations (out of 105), namely
the age (from 15 to 30), age (from 30 to 45), gender, black hair,
and short hair. The selection is based on (1) enough training samples
($>25\%$ of the PETA for both positive/negative classes), (2) upper
body related attributes (3) attributes that conventionally persist
across a day.

For detailed results on the attributes, see Table \ref{tab:validation-set-attributes-accuracy}.
We note that the gender cues (both PIPA and PETA) give the greatest
performance gain.

\clearpage{}

\clearpage{}

\begin{table}
\begin{centering}
\begin{tabular}{llc}
\multicolumn{2}{l}{Cue} & Accuracy\tabularnewline
\hline 
\hline 
\multicolumn{2}{l}{Chance level} & \hspace{0.5em}0.27\tabularnewline
\hline 
Scene & \texttt{$\texttt{s}$} & 27.06\tabularnewline
Body  & $\mbox{\ensuremath{\texttt{b}}}$ & 80.81\tabularnewline
Upper body & $\texttt{u}$ & 84.76\tabularnewline
Head & $\texttt{h}$ & 83.88\tabularnewline
Face & $\texttt{f}$ & 74.45\tabularnewline
\hline 
Zoom out & $\texttt{f}$ & 74.45\tabularnewline
 & $\texttt{f}\negthinspace+\negthinspace\texttt{h}$ & 84.80\tabularnewline
 & $\texttt{f}\negthinspace+\negthinspace\texttt{h}\negthinspace+\negthinspace\texttt{u}$ & 90.65\tabularnewline
 & $\texttt{f}\negthinspace+\negthinspace\texttt{h}\negthinspace+\negthinspace\texttt{u}\negthinspace+\negthinspace\texttt{b}$ & 91.14\tabularnewline
 & $\texttt{f}\negthinspace+\negthinspace\texttt{h}\negthinspace+\negthinspace\texttt{u}\negthinspace+\negthinspace\texttt{b}\negthinspace+\negthinspace\texttt{s}$ & 91.16\tabularnewline
\hline 
Zoom in & \texttt{$\texttt{s}$} & 27.06\tabularnewline
 & $\texttt{s}\negthinspace+\negthinspace\texttt{b}$ & 82.16\tabularnewline
 & $\texttt{s}\negthinspace+\negthinspace\texttt{b}\negthinspace+\negthinspace\texttt{u}$ & 86.39\tabularnewline
 & $\texttt{s}\negthinspace+\negthinspace\texttt{b}\negthinspace+\negthinspace\texttt{u}\negthinspace+\negthinspace\texttt{h}$ & 90.40\tabularnewline
 & $\texttt{s}\negthinspace+\negthinspace\texttt{b}\negthinspace+\negthinspace\texttt{u}\negthinspace+\negthinspace\texttt{h}\negthinspace+\negthinspace\texttt{f}$ & 91.16\tabularnewline
\hline 
Head+body & $\texttt{h}\negthinspace+\negthinspace\texttt{b}$ & 89.42\tabularnewline
Face+head & $\texttt{f}\negthinspace+\negthinspace\texttt{h}$ & 84.80\tabularnewline
 & $\texttt{f}\negthinspace+\negthinspace\texttt{h}\negthinspace+\negthinspace\texttt{u}$ & 90.65\tabularnewline
 & $\texttt{f}\negthinspace+\negthinspace\texttt{h}\negthinspace+\negthinspace\texttt{b}$ & 90.19\tabularnewline
Full person & $\texttt{P}=\texttt{f}\negthinspace+\negthinspace\texttt{h}\negthinspace+\negthinspace\texttt{u}\negthinspace+\negthinspace\texttt{b}$\hspace*{-1.5em} & 91.14\tabularnewline
Full image & $\texttt{\ensuremath{\mbox{\ensuremath{\texttt{P}}}_{s}}}=\texttt{P}\negthinspace+\negthinspace\texttt{s}$ & 91.16\tabularnewline
\end{tabular}
\par\end{centering}

\vspace{0em}

\protect\caption{\label{tab:validation-set-regions-accuracy-appendix}Validation set
accuracy of different cues.}
\end{table}
\begin{table}
\begin{centering}
\begin{tabular}{c|cccc}
\diaghead{\hskip5.1em}{Method}{Setup} & Original & Album & Time & Day\tabularnewline
\hline 
Chance level & 0.17 & 0.17 & 0.17 & 0.50\tabularnewline
$\mbox{\ensuremath{\texttt{h}}}_{\texttt{rgb}}$  & 33.77 & 27.19 & 16.91 & 6.78\tabularnewline
\hline 
$\mbox{\ensuremath{\texttt{s}}}$  & 24.71 & 19.89 & 12.83 & 8.67\tabularnewline
$\mbox{\ensuremath{\texttt{b}}}$  & 69.63 & 59.29 & 44.92 & 20.41\tabularnewline
$\mbox{\ensuremath{\texttt{h}}}$ & 76.42 & 67.48 & 57.05 & 36.37\tabularnewline
$\mbox{\ensuremath{\texttt{h}}}+\mbox{\ensuremath{\texttt{b}}}$ & 83.36 & 73.97 & 63.03 & 38.12\tabularnewline
$\texttt{P}=\texttt{f}\negthinspace+\negthinspace\texttt{h}\negthinspace+\negthinspace\texttt{u}\negthinspace+\negthinspace\texttt{b}$ & 85.33 & 76.49 & 66.55 & 42.14\tabularnewline
$\texttt{\ensuremath{\mbox{\ensuremath{\texttt{P}}}_{s}}}=\texttt{P}\negthinspace+\negthinspace\texttt{s}$ & 85.71 & 76.68 & 66.55 & 42.24\tabularnewline
\texttt{$\texttt{naeil}=\texttt{\ensuremath{\mbox{\ensuremath{\texttt{P}}}_{s}}+\texttt{E}}$} & 86.78 & 78.72 & 69.29 & 46.61\tabularnewline
\hline 
\texttt{PIPER \cite{Zhang2015CvprPiper}} & 83.05 & - & - & -\tabularnewline
\end{tabular}
\par\end{centering}

\vspace{0em}

\protect\caption{\label{tab:splits-accuracy}Recognition accuracy across different
experimental setups on the test data.\protect \\
Extended data $\texttt{E}=\mbox{\ensuremath{\texttt{h}}}_{\texttt{casia}}\negthinspace+\negthinspace\mbox{\ensuremath{\texttt{h}}}_{\texttt{cacd}}\negthinspace+\negthinspace\texttt{\ensuremath{\mbox{\ensuremath{\texttt{h}}}_{\texttt{pipa11}}}+\ensuremath{\mbox{\ensuremath{\texttt{u}}}_{\texttt{peta5}}}}$.}
\end{table}
\begin{table}
\begin{centering}
\begin{tabular}{llc}
 & Method & Accuracy\tabularnewline
\hline 
\hline 
Head & $\texttt{h}$ & 83.88\tabularnewline
 & $\texttt{h}_{\texttt{nn}}$ & 74.92\tabularnewline
\hline 
Head+Body & $\texttt{h}\negthinspace+\negthinspace\texttt{b}$ & 89.42\tabularnewline
 & $\left\{ \texttt{h}\negthinspace+\negthinspace\texttt{b}\right\} _{\texttt{nn}}$ & 79.63\tabularnewline
\hline 
Full Person & $\texttt{P}=\texttt{f}\negthinspace+\negthinspace\texttt{h}\negthinspace+\negthinspace\texttt{u}\negthinspace+\negthinspace\texttt{b}$ & 91.14\tabularnewline
 & $\texttt{P}_{\texttt{nn}}$ & 77.31\tabularnewline
\end{tabular}
\par\end{centering}

\vspace{0em}

\protect\caption{\label{tab:svm-vs-nn}Validation set accuracy using SVM versus NN.}
\end{table}
\begin{table}
\begin{centering}
\begin{tabular}{llc}
Attribute & Classes & Criteria\tabularnewline
\hline 
\hline 
Age & Infant & {\footnotesize{}Not walking (due to young age)}\tabularnewline
 & Child & {\footnotesize{}Not fully grown body size}\tabularnewline
 & Young Adult & {\footnotesize{}Fully grown \& Age $<45$}\tabularnewline
 & Middle Age & {\footnotesize{}$45\leq\mbox{Age}<60$}\tabularnewline
 & Senior & {\footnotesize{}Age$\geq60$}\tabularnewline
\hline 
Gender & Female & {\footnotesize{}Female looking}\tabularnewline
 & Male & {\footnotesize{}Male looking}\tabularnewline
\hline 
Glasses & None & {\footnotesize{}No eyewear}\tabularnewline
 & Glasses & {\footnotesize{}Glasses without eye occlusion}\tabularnewline
 & Sunglasses & {\footnotesize{}Glasses with eye occlusion}\tabularnewline
\hline 
Haircolour & Black & {\footnotesize{}Black}\tabularnewline
 & White & {\footnotesize{}Any hint of whiteness}\tabularnewline
 & Others & {\footnotesize{}Neither of the above}\tabularnewline
\hline 
Hairlength & No hair & {\footnotesize{}No hair on the scalp}\tabularnewline
 & Less hair & {\footnotesize{}Hairless for $>\frac{1}{2}$ upper scalp}\tabularnewline
 & Short hair & {\footnotesize{}When straightened,$<10$ cm}\tabularnewline
 & Med hair & {\footnotesize{}When straightened, $<$chin level}\tabularnewline
 & Long hair & {\footnotesize{}When straightened, $>$chin level}\tabularnewline
\end{tabular}
\par\end{centering}

\begin{centering}

\par\end{centering}

\vspace{0em}

\protect\caption{\label{tab:attributes-details}PIPA attributes details.}
\end{table}
\begin{table}
\begin{centering}
\begin{tabular}{llc}
 & Method & Accuracy\tabularnewline
\hline 
\hline 
Head & $\texttt{h}$ & 83.88\tabularnewline
\hline 
 & $\mbox{\ensuremath{\texttt{h}}}_{\texttt{pipa11m}}$  & 74.63\tabularnewline
 & $\mbox{\ensuremath{\texttt{h}}}_{\texttt{pipa11}}$  & 81.74\tabularnewline
 & $\texttt{h}+\mbox{\ensuremath{\texttt{h}}}_{\texttt{pipa11}}$ & 85.00\tabularnewline
\hline 
 & $\texttt{h}+\mbox{\ensuremath{\texttt{h}}}_{\texttt{age}}$ & 84.40\tabularnewline
 & $\texttt{h}+\mbox{\ensuremath{\texttt{h}}}_{\texttt{gender}}$ & \textbf{84.69}\tabularnewline
 & $\texttt{h}+\mbox{\ensuremath{\texttt{h}}}_{\texttt{glasses}}$ & 84.30\tabularnewline
 & $\texttt{h}+\mbox{\ensuremath{\texttt{h}}}_{\texttt{haircolour}}$ & 84.25\tabularnewline
 & $\texttt{h}+\mbox{\ensuremath{\texttt{h}}}_{\texttt{hairlength}}$ & 84.39\tabularnewline
\hline 
Upper Body & $\texttt{u}$ & 84.76\tabularnewline
\hline 
 & $\mbox{\ensuremath{\texttt{u}}}_{\texttt{peta5m}}$ & 75.71\tabularnewline
 & $\mbox{\ensuremath{\texttt{u}}}_{\texttt{peta5}}$ & 77.50\tabularnewline
 & $\texttt{u}+\mbox{\ensuremath{\texttt{u}}}_{\texttt{peta5}}$ & 85.18\tabularnewline
\hline 
 & $\texttt{u}+\mbox{\ensuremath{\texttt{u}}}_{\texttt{age1}}$ & 84.75\tabularnewline
 & $\texttt{u}+\mbox{\ensuremath{\texttt{u}}}_{\texttt{age2}}$ & 84.81\tabularnewline
 & $\texttt{u}+\mbox{\ensuremath{\texttt{u}}}_{\texttt{gender}}$ & \textbf{84.90}\tabularnewline
 & $\texttt{u}+\mbox{\ensuremath{\texttt{u}}}_{\texttt{hairshort}}$ & 84.87\tabularnewline
 & $\texttt{u}+\mbox{\ensuremath{\texttt{u}}}_{\texttt{hairblack}}$ & 84.80\tabularnewline
\hline 
Head+Upper Body & $\mbox{\ensuremath{\texttt{A}}}=\mbox{\ensuremath{\texttt{h}}}_{\texttt{pipa11}}+\mbox{\ensuremath{\texttt{u}}}_{\texttt{peta5}}$\hspace*{-1.5em} & 86.17\tabularnewline
 & $\texttt{h}+\texttt{u}$ & 85.77\tabularnewline
 & $\texttt{h}+\texttt{u}+\ensuremath{\texttt{A}}$ & 90.12\tabularnewline
\end{tabular}
\par\end{centering}

\vspace{0em}

\protect\caption{\label{tab:validation-set-attributes-accuracy}Validation set accuracy
of different attribute cues.}
\end{table}

\clearpage{}

\clearpage{}

\end{document}